\documentclass{article}

\usepackage{arxiv}

\usepackage[utf8]{inputenc} 
\usepackage[T1]{fontenc}    
\usepackage{hyperref}       
\usepackage{url}            
\usepackage{booktabs}       
\usepackage{amsfonts}       
\usepackage{nicefrac}       
\usepackage{microtype}      
\usepackage{lipsum}
\usepackage{comment}
\usepackage{graphicx}
\usepackage{amsmath}
\usepackage{amsthm}
\usepackage{booktabs}
\usepackage{algorithm}
\usepackage{url}
\usepackage{algorithmic}
\usepackage[switch]{lineno}
\usepackage{natbib}
\usepackage{multirow}
\usepackage{amsmath,amssymb,amsfonts,amsthm}
\usepackage{caption}
\usepackage{subcaption}
\usepackage{booktabs}
\newtheorem{example}{Example}
\newtheorem{theorem}{Theorem}
\usepackage{dsfont}
\newtheorem{definition}{Definition}
\newtheorem{assumption}{Assumption}
\usepackage{adjustbox}

\usepackage{graphicx}
\graphicspath{ {./images/} }

\title{Provably Robust Bayesian Counterfactual Explanations under Model Changes}

\author{
 Jamie Duell \\
  School of Computing and Digital Technologies\\
  Sheffield Hallam University \\
  \texttt{j.duell@shu.ac.uk} \\
   \And
 Xiuyi Fan\\
  Lee Kong Chian School of Medicine\\
  Nanyang Technological University \\
  \texttt{xyfan@ntu.edu.sg}  \\
}

\begin{document}
\maketitle
\begin{abstract}
Counterfactual explanations (CEs) offer interpretable insights into machine learning predictions by answering ``what if?" questions. However, in real-world settings where models are frequently updated, existing counterfactual explanations can quickly become invalid or unreliable. In this paper, we introduce Probabilistically Safe CEs (PSCE), a method for generating counterfactual explanations that are $\delta$-safe, to ensure high predictive confidence, and $\epsilon$-robust to ensure low predictive variance. Based on Bayesian principles, PSCE provides formal probabilistic guarantees for CEs under model changes which are adhered to in what we refer to as the $\langle \delta, \epsilon \rangle$-set. Uncertainty-aware constraints are integrated into our optimization framework and we validate our method empirically across diverse datasets. We compare our approach against state-of-the-art Bayesian CE methods, where PSCE produces counterfactual explanations that are not only more plausible and discriminative, but also provably robust under model change.
\end{abstract}

\section{Introduction}

Counterfactual Explanations (CEs) have gained traction as a way to answer "what-if?" questions in machine learning (ML) \citep{wachter}. A CE typically seeks a minimal alteration to an input that changes the model's prediction to a desired class. To be reliable, CEs should satisfy core desiderata such as \emph{proximity}, \emph{validity}, \emph{discriminativeness}, \emph{robustness} and \emph{plausibility} \citep{sokol2025needcounterfactualexplainabilityprincipled}. However, a critical challenge often overlooked is the robustness of CEs in dynamic environments. When an underlying ML model is updated, which is a common scenario in real-world applications such as online and continual learning \citep{hoi2018onlinelearningcomprehensivesurvey, vandeVen2022} explanations, the generated for the original model may become invalid. Small shifts in model parameters can move decision boundaries, rendering previous CEs useless.

While some recent work has proposed post-hoc methods to update CEs after a model changes \citep{DBLP:conf/icml/HammanNMMD23, 10.1145/3690624.3709300}, a more principled approach is to generate explanations that are inherently robust to such perturbations from the outset. This requires integrating model uncertainty directly into the CE generation process. By considering a distribution over model parameters, as is natural in Bayesian frameworks like Bayesian Neural Networks (BNNs) \citep{10.5555/3454287.3455600} or Monte Carlo Dropout (MC Dropout) \citep{pmlr-v48-gal16}, we can search for CEs that are not just valid for a single point estimate of the model, but remain valid across a whole family of plausible models.

The core intuition is that a robust CE should be classified with both 1) high predictive certainty (a high mean probability for the target class) and 2) low predictive variance across the model posterior. A CE with these properties resides deep within the target class's decision region, creating a "safety buffer" that makes it resilient to minor shifts in the decision boundary caused by model changes.
\begin{figure}[h]
\centering
\includegraphics[width=0.75\linewidth]{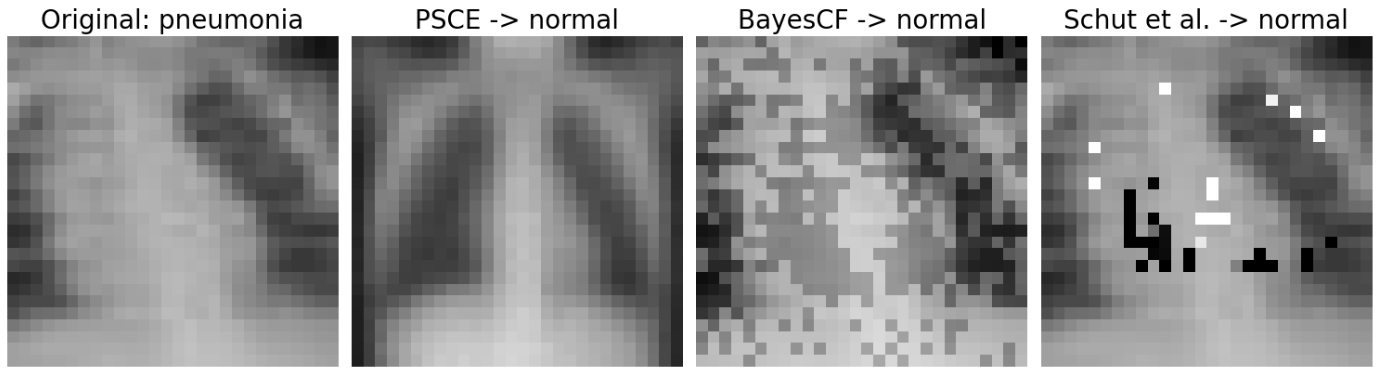}
\caption{Counterfactual examples for a PneumoniaMNIST \citep{medmnistv1, medmnistv2} image, transitioning from `pneumonia'(on the left) to `normal'. The output from our proposed method, PSCE, is shown alongside other Bayesian approaches discussed in this work.}
\label{fig:pneumnist}
\end{figure}
To formalize this, we introduce Probabilistically Safe Counterfactual Explanations (PSCE), a Bayesian-inspired method that generates CEs with formal guarantees on their robustness. PSCE optimizes for counterfactuals that are not only robust but also plausible by encouraging them to lie on the data manifold. This results in coherent and interpretable outputs, as illustrated in the medical imaging example in Figure \ref{fig:pneumnist}.

\begin{figure}[H]
\centering
\includegraphics[width=0.75\linewidth]{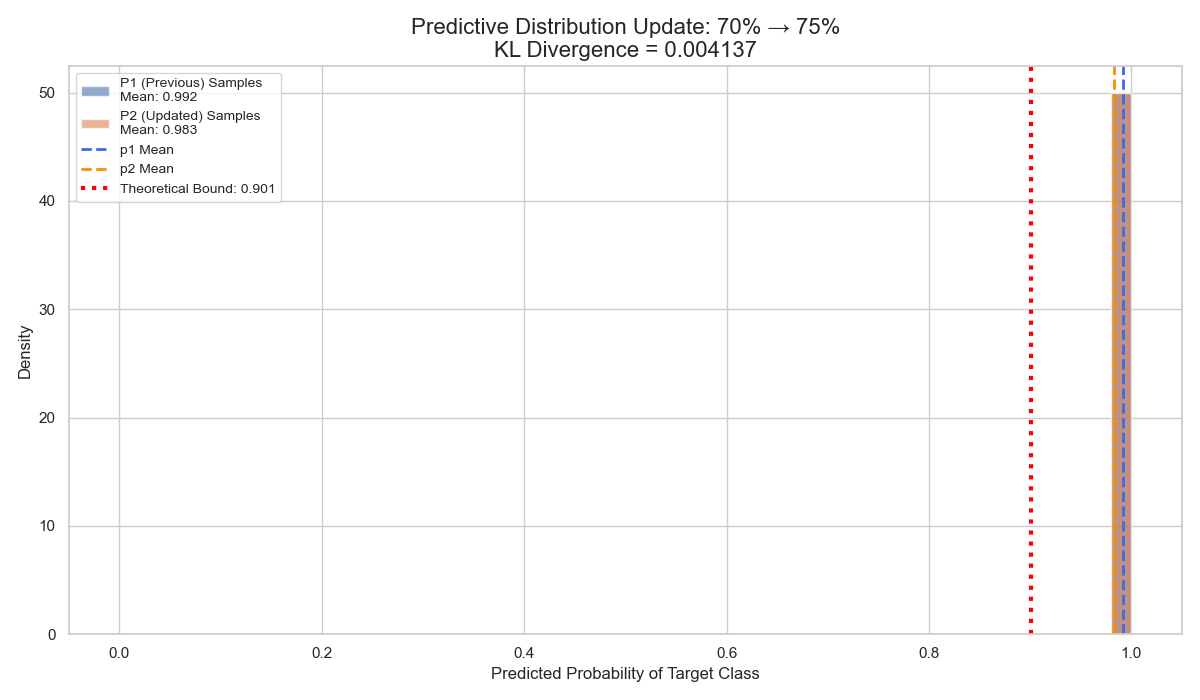}
\caption{Predictive distribution for a CE before (p1) and after (p2) a 5\% data increment. The PSCE-generated CE remains confidently classified, and the new prediction is well-described by our theoretical lower bound.}
\label{fig:dist_shift}
\end{figure}

Beyond generating plausible explanations, PSCE's primary guarantee is robustness to model changes. We illustrate this core property with a simple example:

\begin{example}\label{example_2}
Suppose a counterfactual is generated for a model trained on 70\% of the available data. The model is then updated by training on an additional 5\% of the data, causing its posterior distribution to shift. As shown in Figure \ref{fig:dist_shift}, the predictive probability for the PSCE-generated counterfactual remains well above the required threshold, demonstrating its robustness to the model change. Our theoretical bound, introduced later, formally guarantees this outcome.
\end{example}

In this work, we compare PSCE against other Bayesian, model-specific CE methods: BayesCF \citep{batten2025uncertaintyaware} and Schut et al. \citep{Schut2021GeneratingIC}. Our central contributions are threefold:
\begin{enumerate}
\item We introduce PSCE, a method for generating probabilistically safe counterfactuals that explicitly optimizes for high confidence and low variance.
\item We derive a theoretical bound that formally guarantees the validity of a counterfactual under a given magnitude of model change. 
\item We empirically validate the robustness of PSCE against existing Bayesian CE methods, demonstrating its superior performance across key metrics from counterfactual literature.
\end{enumerate}

\section{Counterfactual Desiderata}

In this section, we informally discuss counterfactual desiderata as per recent literature \citep{sokol2025needcounterfactualexplainabilityprincipled}, ensuring that our approach is principled. Thus, we have the following: 
\begin{itemize}
    \item \textbf{Validity:} Ensures the counterfactual instance falls into a desired counterfactual class.
    \item \textbf{Robustness:} Ensures that sufficiently small model changes do not invalidate counterfactual instances from being classified under the desired counterfactual class. 
    \item \textbf{Plausibility:} Ensures that counterfactual instance is on the data manifold and thus is plausible as per the data.
    \item \textbf{Discriminativeness:} Ensures that the counterfactual is distinguishable, thus looks like it belongs to the counterfactual class.
    \item \textbf{Proximity:} Ensures that minimal changes are needed across all features to obtain a desired counterfactual instance.
\end{itemize}

\section{Robust and Probabilistically Safe Counterfactuals}

To begin with the building blocks that constitute the central objective function, we consider a Bayesian lens for classification, in which the model parameters $\omega$ are treated as random variables with a posterior distribution given the data $\mathcal{D}$. This takes the form: 
\begin{align*}
    p(\omega \vert \mathcal{D}) = \frac{p(\mathcal{D}\vert \omega)p(\omega)}{p(\mathcal{D})},
\end{align*}
where $p(\omega \vert \mathcal{D})$ is the posterior distribution of the model parameters given the training data $\mathcal{D}$. In this work, we use both BNNs \citep{10.5555/3454287.3455600} and  MC Dropout \citep{pmlr-v48-gal16} to approximate $p(\omega \vert \mathcal{D})$. From a Bayesian perspective, we begin by defining a classifier:
\begin{definition}[Bayesian Classifier]
    A classifier with parameters $\omega$ maps an input $\mathbf{x} \in \mathbf{X}$ to logits: 
\begin{align*}
\mathbf{l}(\mathbf{x}; \omega) = \langle l_1(\mathbf{x}; \omega), \ldots, l_C(\mathbf{x}; \omega)\rangle \in \mathbb{R}^C,
\end{align*}
the class probabilities are obtained via the softmax transformation: 
\begin{align*}
f_c(\mathbf{x}; \omega) = \frac{\exp(l_c(\mathbf{x}; \omega))}{\sum_{i=1}^C \exp(l_i(\mathbf{x}; \omega))}, \quad c=1,\ldots,C.
\end{align*}
Under the Bayesian formulation, the posterior predictive distribution is given by:  
\begin{align*}
    p(y =c \vert \mathbf{x}, \mathcal{D}) = \mathbb{E}_{\omega \sim p(\omega \vert \mathcal{D})}\bigg[f_c(\mathbf{x}; \omega)  \bigg], 
\end{align*}
the predicted label is: 
\begin{align*}
    y = \underset{c \in \{1,\ldots, C\} }{\arg \max} \ p(y =c \vert \mathbf{x}, \mathcal{D}).
\end{align*}
\end{definition}
Given a classifier, we can then define a counterfactual instance from a Bayesian perspective to be given:
\begin{definition}[Counterfactual Instance]
Let $\mathbf{x} \in \mathbb{R}^J$ be an instance, $\mathcal{G}_{CF}: \mathbb{R}^J \to \mathbb{R}^J$ a counterfactual generator, and $y$ the predicted class of $\mathbf{x}$ under the Bayesian classifier. A counterfactual instance $\mathbf{x}^\prime$ produced by a counterfactual generator $\mathcal{G}_{CF}$ satisfies: 
\begin{align*}
\mathbf{x}^\prime :  \underset{c \in \{1,\ldots,C\}}{\operatorname{arg\,max}} \ p(y^\prime = c \vert \mathbf{x}^\prime, \mathcal{D}) \neq y \\ \text{where, } \mathbf{x}^\prime = \mathcal{G}_{CF}(\mathbf{x}).
\end{align*}
where, $y^\prime$ is the counterfactual class. 
\end{definition}

A posterior predictive distribution for a counterfactual $\mathbf{x}^\prime$ belonging to a counterfactual class $y^\prime$ is given as: 
\begin{align*}
    p(y^\prime \vert \mathbf{x}^\prime, \mathcal{D}) &= \int p(y^\prime \vert \mathbf{x}^\prime, \omega) p(\omega \vert \mathcal{D}) d \omega = \mathbb{E}_{\omega \sim p(\omega \vert \mathcal{D})}\bigg[ p(y^\prime \vert \mathbf{x}^\prime, \omega) \bigg].
\end{align*}

To ensure counterfactuals remain valid, we additionally require that they lie in high-likelihood regions of the data distribution, which we model via a generative density $p_\theta(\mathbf{x})$ in the model. The likelihood of a candidate counterfactual instance $\mathbf{x}^\prime$ being valid can be assessed through the learned likelihood $p_\theta(\mathbf{x})$. Parameters $\theta$ are optimised for the dataset $\mathbf{X}$ according to:
\begin{align*}
\theta^* &:= \underset{\theta}{\arg\max} \ \frac{1}{|\mathbf{X}|}\sum_{\mathbf{x} \in \mathbf{X}} \log p_{\theta}(\mathbf{x}) = \underset{\theta}{\arg\max} \ \frac{1}{|\mathbf{X}|}\sum_{\mathbf{x} \in \mathbf{X}} \log \int p_{\theta}(\mathbf{x} \vert \mathbf{z})p(\mathbf{z})  d \mathbf{z}.
\end{align*}
This corresponds to maximising the marginal log-likelihood of the data. In doing so, $p_\theta$ approximates the true data distribution $\mathcal{D}$, such that any new counterfactual instance $\mathbf{x}^\prime \sim \mathcal{D}$ is expected to have high likelihood under the distribution $p_\theta$.

As suggested, a common occurrence in real-world deployment is \emph{model change}. Existing works discuss robustness to model change \citep{10.24963/ijcai.2024/894}, which informally states that a counterfactual remains valid under a ``sufficiently small" change in model parameters. Our strategy, is to propose a tractable optimization scheme to quantify robustness under model changes. The intuition is that a counterfactual instance which is confidently embedded within the target class's decision region for some original model is more likely to survive minor shifts in the decision boundary from a model change, this sentiment is shared in the work \citep{DBLP:conf/icml/HammanNMMD23}. We can formalize this by ensuring that two properties affiliated with the posterior $p(\omega \vert \mathcal{D})$ are encouraged, namely: 
\begin{itemize}
    \item High Predictive Certainty: The model should, on average, classify the counterfactual with very high confidence.
    \item Low Predictive Variance: The prediction should be stable across all plausible versions of the model described by the posterior. High variance would imply that some versions of the model already place the counterfactual near or over a decision boundary, thus making it fragile.
\end{itemize}

These two properties give rise to our building blocks: $\delta$-safety and 
$\epsilon$-robustness. We later provide a theoretical bound on the robustness and safety of a counterfactual under model changes under an updated posterior which cannot be known before hand. 
 
To begin, we say a counterfactual is $\delta$-safe, if the posterior predictive distribution falls safely over a decision threshold $1-\delta$ for $\delta \in [0,1]$. More formally:  
\begin{definition}[$\delta$-safe Counterfactual]\label{delta_safe}
Given a counterfactual class $y^\prime$ and a data distribution $\mathcal{D}$, a counterfactual $\mathbf{x}^\prime$ is considered $\delta$-safe, if and only if:
\begin{align*}
p(y^\prime \vert \mathbf{x}^\prime, \mathcal{D}) \geq 1 - \delta.
\end{align*}

\end{definition}

Extending this, we consider a counterfactual to be $\epsilon$-robust, if the variance of the likelihood falls under some threshold $\epsilon$. Formally,
\begin{definition}[$\epsilon$-robust Counterfactual]\label{eps_safe}
Given a counterfactual class $y^\prime$, a data distribution $\mathcal{D}$ and a distribution defined over model weights given the data distribution $p(\omega\vert \mathcal{D})$, we consider a counterfactual $\mathbf{x}^\prime$ to be a $\epsilon$-robust if and only if: 
\begin{align*}
    \text{Var}_{\omega \sim p(\omega \vert \mathcal{D})}\bigg[p(y^\prime \vert \mathbf{x}^\prime, \omega)  \bigg] \leq \epsilon. 
\end{align*}
\end{definition}
We take the approach, as this represents epistemic uncertainty which is reducible by definition, and thus optimizing variance under some threshold $\epsilon$ is a plausible approach to reduce uncertainty for a counterfactual instance $\mathbf{x}^\prime$.

Finally, we can then prescribe rules for what we name the $\langle\delta,\epsilon\rangle$-set of counterfactual instances. 
 \begin{definition}[$\langle\delta,\epsilon\rangle$-Set]
     Counterfactual instances $\mathbf{x}^\prime$ belong to the $\langle\delta,\epsilon\rangle$-set ($S_{\langle\delta,\epsilon\rangle}$), where:
     \begin{align*}
         S_{\langle\delta,\epsilon\rangle} := \ \{\mathbf{x}^\prime &: p(y^\prime \vert \mathbf{x}^\prime, \mathcal{D}) \geq 1-\delta \ \wedge     \text{Var}_{\omega \sim p(\omega \vert \mathcal{D})}\bigg[p(y^\prime \vert \mathbf{x}^\prime, \omega)  \bigg] \leq \epsilon. \} ,
     \end{align*}
     where the counterfactual instance $\mathbf{x}^\prime$ is both $\delta$-safe, $\epsilon$-robust. 
 \end{definition}
The definition for the $\langle\delta,\epsilon\rangle$-set, namely $S_{\langle\delta,\epsilon\rangle}$ ensures safety guarantees that consider both the standard deviation and the posterior predictive distribution. By filtering a set of counterfactual instances $S_{\langle\delta,\epsilon\rangle}$, we can ensure robustness to model changes, where we provide provable theoretical guarantees introduced in our central theorems, namely Theorem 1 and Theorem 2. We then proceed to empirically verify the utility of our central theorems.

\section{Generating Counterfactuals}
To achieve this, we ensure that a counterfactual is $\delta$-safe, $\epsilon$-robust and adheres to desiderata presented in the previous section. To this end, we propose the following optimization function to generate a robust counterfactual $\mathbf{x}^\prime$, initialized at $\mathbf{x}$:
\begin{align*}
  \mathbf{x}^\prime := \mathcal{G}_{PSCE}(\mathbf{x}) = \underset{\mathbf{x}^\prime}{\arg \min} \ &(\lambda_1 \mathcal{L}_{clf} + \lambda_2 \mathcal{L}_{del} + \lambda_3 \mathcal{L}_{ldist} + \lambda_4 \mathcal{L}_{var} - \lambda_5 \mathcal{L}_{ELBO}).
\end{align*}
From this optimization problem, we define each part independently. Thus, we have:
\begin{align*}
    \mathcal{L}_{clf}(\mathbf{x}^\prime) = - \mathbb{E}_{\omega \sim p(\omega \vert \mathcal{D})}[\text{log } p(y^\prime \vert \mathbf{x}^\prime,\omega)]. 
\end{align*}
The $\mathcal{L}_{clf}$ term aims to ensure that the generated counterfactual $\mathbf{x}^\prime$ belongs to a counterfactual class $y^\prime$, which corresponds to the negative expected log-likelihood over the posterior distribution of weights $\omega$. We then introduce:
\begin{align*}
    \mathcal{L}_{del}(\mathbf{x}^\prime) = \text{max}((1-\delta)-\mathbb{E}_{\omega \sim p(\omega \vert \mathcal{D})}[ p(y^\prime \vert \mathbf{x}^\prime,\omega)],0), 
\end{align*}
which enforces that the predictive probability for $\mathbf{x}^\prime$ is at least $1-\delta$, consistent with the definition of a $\delta$-safe counterfactual. Under similar notation, we have the loss function aligning with an $\epsilon$-robust counterfactual, given by:  
\begin{align*} 
    \mathcal{L}_{var}(\mathbf{x}^\prime) =    \text{max}(\text{Var}_{\omega \sim p(\omega\vert \mathcal{D})}[p(y^\prime \vert \mathbf{x}^\prime, \omega)] - \epsilon,0).
\end{align*}

Next, we consider plausibility. In the context of our work, plausibility refers to the likelihood of observing a counterfactual $\mathbf{x}^\prime$ under the data distribution $\mathcal{D}$, parameterized by $\theta^\prime$. In practice, one does not know the true distribution $\theta^\prime$, and thus optimization over some parameter $\theta$ to obtain an optimal $\theta^*$ such that $\theta^\prime \approx \theta^*$ is computed. Thus, by equation \ref{prob_x}, we follow the standard Variational Autoencoder (VAE) \citep{DBLP:journals/corr/KingmaW13} formulation, introducing the approximate posterior $q_{\phi}(\mathbf{z} \vert \mathbf{x})$ such that:
\begin{align}\label{prob_x}
    \text{log }p(\mathbf{x}) = \text{log }\int q_{\phi}(\mathbf{z} \vert \mathbf{x})\frac{p_{\theta}(\mathbf{x}\vert\mathbf{z})p(\mathbf{z})}{q_{\phi}(\mathbf{z} \vert \mathbf{x})} d \mathbf{z}. 
\end{align} 
Assuming parameters $\{ \theta^*, \phi^* \}$ are learned for the pretrained model. Then, we can adopt the standard Evidence Lower Bounds (ELBO) to ensure that a counterfactual $\mathbf{x}^\prime$ remains within the data distribution.
\begin{align*}
    \mathcal{L}_{ELBO}(\mathbf{x}^\prime) = & \mathbb{E}_{q_{\phi^*}(\mathbf{z} \vert \mathbf{x}^\prime)}[\text{log }p_{\theta^*}(\mathbf{x}^\prime\vert \mathbf{z})] + \mathbb{E}_{q_{\phi^*}(\mathbf{z} \vert \mathbf{x}^\prime)}[\text{log }p(\mathbf{z})] - \mathbb{E}_{q_{\phi^*}(\mathbf{z} \vert \mathbf{x}^\prime)}[\text{log }q_{\phi^*}(\mathbf{z} \vert \mathbf{x}^\prime)],
\end{align*}
where $\theta^*$ and $\phi^*$ are given by a pretrained variational autoencoder (VAE), the standard ELBO derivation is provided in the \emph{supplementary material}. All the expectations in this section are approximated through Monte Carlo sampling with dropout enabled at the time of inference or sampling over BNN weights.

To account for proximity, we adopt the technique of minimizing distance of generated samples within a latent space. Thus, given an instance $\mathbf{x}$ and its counterfactual $\mathbf{x}^\prime$, let $\mu_{\phi^*}(\mathbf{x})$ represent the mean encoded representation of $\mathbf{x}$, and $\mu_{\phi^*}(\mathbf{x}^\prime)$ be the mean encoded representation of $\mathbf{x}^\prime$, then the final term of our objective function is given as: 
\begin{align*}
    \mathcal{L}_{ldist}(\mathbf{x}^\prime, \mathbf{x}) = &\vert \vert \mu_{\phi^*}(\mathbf{x}) - \mu_{\phi^*}(\mathbf{x}^\prime)\vert \vert^2_2.
\end{align*}
Informally, this encourages the encoded counterfactual to remain close to the encoded representation of the original instance. In principle, our approach provides optimization in accordance to the following desiderata: 
\begin{itemize}
    \item $\mathcal{L}_{clf}$ ensures \emph{validity} of the counterfactual, whereby $\mathcal{L}_{del}$ ensures the \emph{validity} is safe under a distribution of parameters $p(\omega\vert \mathcal{D})$ -- which further inclines the safety of counterfactual explanations under model changes which we address in the next section (concerning \emph{robustness}).
    \item $\mathcal{L}_{var}$ ensures that the counterfactual is \emph{validity} by encouraging epistemic uncertainty in counterfactual generation to be less than or equal to some $\epsilon$. 
    \item $\mathcal{L}_{ELBO}$ and $\mathcal{L}_{ldist}$ aim to ensure that 1) the counterfactual pathway is optimised such that it remains within the data distribution, encouraging \emph{plausibility} and \emph{discriminativeness}, while 2) being in close \emph{proximity} to the origin.
\end{itemize}
More broadly, $\mathcal{L}_{del}$ and $\mathcal{L}_{var}$ focus on the certainty of a counterfactual instance, $\mathcal{L}_{ELBO}$ ensures that the counterfactual is plausible under the learned data distribution, $\mathcal{L}_{ldist}$ encourages similarity, and $\mathcal{L}_{clf}$ directly guides the counterfactual towards the desired counterfactual class.

\section{Counterfactual Robustness under model changes}

In this section, we present Theorem \ref{main_lemma_sup} and \ref{variance_upper_bound}, where the implication is that any counterfactual in the set $S_{\langle \delta, \epsilon \rangle}$ which is encouraged by PSCE has theoretical guarantees on the safety and robustness of a counterfactual instance under model changes. Thus, the theoretical analysis in this section follows assumption 1. 
\begin{assumption}
Counterfactual instances $\mathbf{x}^\prime$ belong to $S_{\langle \delta, \epsilon \rangle}$, the $\langle \delta, \epsilon \rangle$-set. 
\end{assumption}
We first look at the safety of counterfactual instances under model changes, we begin by considering a single update to model parameters from some original data $\mathcal{D}_{prev}$ and some new data $\mathcal{D}_{new}$, whereby a predictive posterior is given as:
\begin{align*}
    p(y^\prime \vert \mathbf{x}^\prime, \mathcal{D}_{prev}) &= \int p(y^\prime \vert \mathbf{x}^\prime,\omega) \cdot p_1(\omega \vert \mathcal{D}_{prev}) d\omega = \mathbb{E}_{\omega \sim p_1(\omega \vert \mathcal{D}_{prev})}[p(y^\prime \vert \mathbf{x}^\prime,\omega)],
\end{align*}
under the original model $\mathcal{D}_{prev}$, and the predictive posterior distribution of the new model is defined as: 
\begin{align*}
    p(y^\prime \vert \mathbf{x}^\prime, &\mathcal{D}_{prev} \cup \mathcal{D}_{new}) = \int p(y^\prime \vert \mathbf{x}^\prime,\omega) \cdot p_2(\omega \vert \mathcal{D}_{prev} \cup \mathcal{D}_{new}) d\omega 
     = \mathbb{E}_{\omega \sim p_2(\omega \vert \mathcal{D}_{prev} \cup \mathcal{D}_{new})}[p(y^\prime \vert \mathbf{x}^\prime,\omega)].
\end{align*}
Here, the posterior predictive distribution is eligible to change as a consequence of a change in the posterior distribution $p(\omega\vert \cdot)$. In which we consider a distribution over the previous data $\mathcal{D}_{prev}$ and the distribution over new data $\mathcal{D}_{new}$, we can then prescribe a posterior distribution over model parameters under the old data (via. $p_1(\omega \vert \cdot)$), and new data (via $p_2(\omega \vert \cdot)$), giving  
\begin{align*}
    p_1(\omega \vert \mathcal{D}_{prev}) = \frac{p(\mathcal{D}_{prev}\vert \omega)p(\omega)}{p(\mathcal{D}_{prev})},
\end{align*}
and
\begin{align*}
    p_2(\omega \vert \mathcal{D}_{prev} \cup \mathcal{D}_{new}) = \frac{p(\mathcal{D}_{prev} \cup \mathcal{D}_{new}\vert \omega)p(\omega)}{p(\mathcal{D}_{prev} \cup \mathcal{D}_{new})}.
\end{align*} 
Assuming that model changes produce small change providing a new posterior distribution $ p_2(\omega \vert \mathcal{D}_{prev} \cup \mathcal{D}_{new})$, that is, the Kullback-Leilber divergence between the old and new posterior is ``sufficiently small" \citep{10.24963/ijcai.2024/894} $D_{\mathrm{KL}}(p_2(\omega \vert \cdot) \vert\vert p_1(\omega \vert \cdot))$ (we explore this in the results section and the \emph{supplementary material}). Theorem \ref{main_lemma_sup} illustrates a tight lower bound for the change in probability for a counterfactual instance $\mathbf{x}^\prime$ under the newly defined posterior distribution. 
\begin{table*}[h]
\centering
\caption{Empirical example for the theoretical bound. A valid counterfactual $\mathbf{x}^\prime$ is selected such that $p_1(y^\prime \vert \mathbf{x}^\prime, \mathcal{D}_{prev}) \geq 1 - \delta$, with $\delta = 0.05$ and a learning rate of $1e-5$. The example uses the Wisconsin Breast Cancer dataset with a Bayesian Neural Network (BNN) to approximate $p_2(\omega \vert \cdot)$ and $p_1(\omega \vert \cdot)$.}
\label{tab:incremental_results}
\begin{tabular}{lccccc}
\toprule
\textbf{Update} & \textbf{$p_1(y^\prime \vert \mathbf{x}^\prime)$} & \textbf{$p_2(y^\prime \vert \mathbf{x}^\prime)$} & $\sim$\textbf{$D_{\mathrm{KL}}(p_2(\omega \vert \cdot) \Vert p_1(\omega \vert \cdot))$} & \textbf{Bound} & \textbf{Holds} \\
\midrule
95\% $\rightarrow$ 96\%  & 0.9977 & 0.9981 & 0.000330 & 0.9720 & \checkmark \\
96\% $\rightarrow$ 97\%  & 0.9981 & 0.9971 & 0.000355 & 0.9714 & \checkmark \\
97\% $\rightarrow$ 98\%  & 0.9971 & 0.9976 & 0.000096 & 0.9832 & \checkmark \\
98\% $\rightarrow$ 99\%  & 0.9976 & 0.9998 & 0.000423 & 0.9685 & \checkmark \\
99\% $\rightarrow$ 100\% & 0.9998 & 0.9974 & 0.000477 & 0.9689 & \checkmark \\
\bottomrule
\end{tabular}
\end{table*}
\begin{figure}
    \centering
    \includegraphics[width=0.65\linewidth]{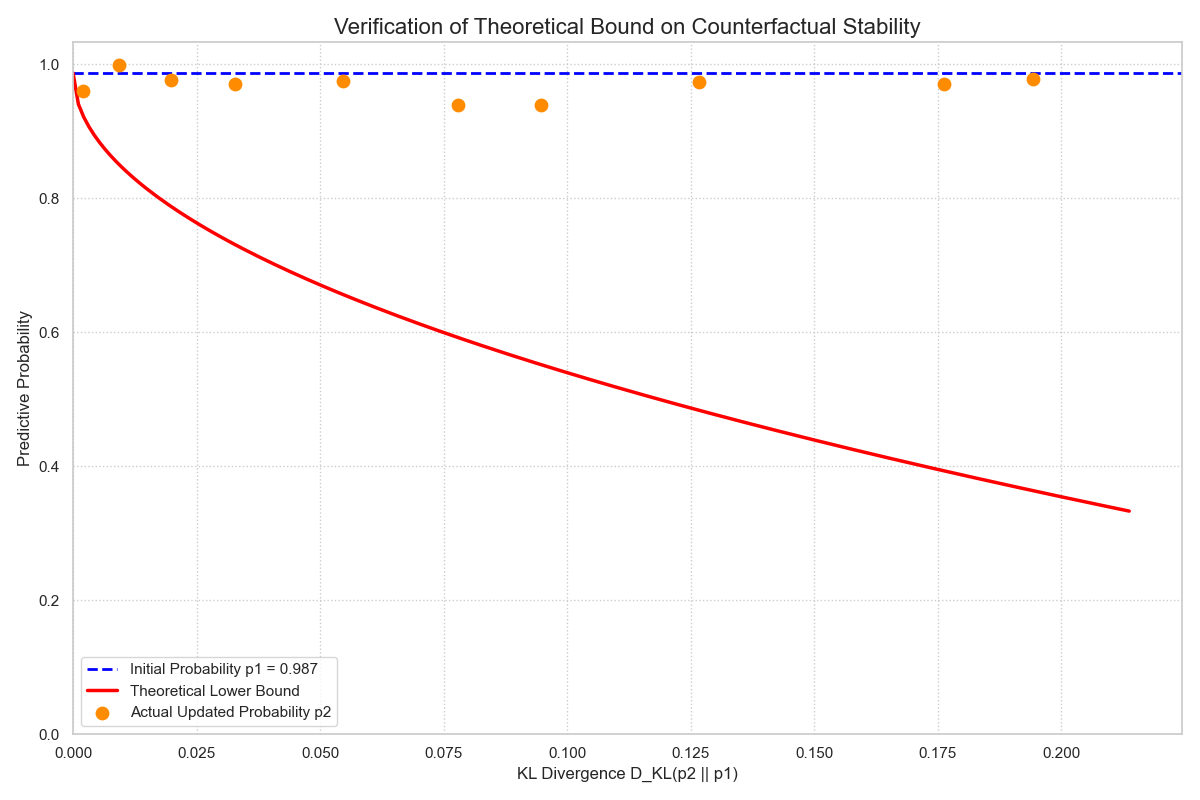}
    \caption{Verification of a closed form solution to the $D_{\mathrm{KL}}$ term presented in the inequality of Theorem 1 associated with the worked example in equation \ref{ineq_closed2}.}
    \label{fig:closedform}
\end{figure}

\begin{theorem}\label{main_lemma_sup}
Let $p_1(\omega \vert \mathcal{D}_{\text{prev}})$ be a probabilistic model's posterior distribution over its parameters $\omega$. After a data update $\mathcal{D}_{\text{prev}} \cup \mathcal{D}_{\text{new}}$, the new posterior is $p_2(\omega \vert \mathcal{D}_{\text{prev}} \cup \mathcal{D}_{\text{new}})$. Given the initial posterior predictive probability for a class $y^\prime$ given a counterfactual $\mathbf{x}^\prime$ is at least $p_1(y^\prime \vert \mathbf{x}^\prime, \mathcal{D}_{prev}) \geq 1 - \delta$, then the updated posterior predictive probability is bounded as:
\begin{align*}
&p_2(y^\prime \vert \mathbf{x}^\prime, \mathcal{D}_{prev} \cup \mathcal{D}_{new}) \\ &\geq (1 - \delta) - 2\cdot \sqrt{\frac{1}{2}D_{\mathrm{KL}}(p_2(\omega \vert \cdot) \vert \vert p_1(\omega \vert \cdot))},
\end{align*}
where, $p_2(\omega \vert \cdot) = p_2(\omega \vert \mathcal{D}_{prev} \cup \mathcal{D}_{new})$ and $p_1(\omega \vert \cdot) = p_1(\omega \vert \mathcal{D}_{prev})$.
\end{theorem}

Thus, we have established that the predictive confidence degrades gracefully under model changes at a rate of $2 \cdot \sqrt{\frac{1}{2}D_{\mathrm{KL}}(p_2(\omega \vert \cdot) \vert \vert p_1(\omega \vert \cdot))}$ relative to the original threshold $1 - \delta$. Thus, a tighter update in the posterior distribution of model parameters under new data $\mathcal{D}_{new}$, the tighter the bound for a counterfactual under model changes. Consequently, we can plug in values directly and solve for $D_{\mathrm{KL}}$, for example, we can consider a scenario where we have a $\delta-$safe (for $\delta = 0.05$) counterfactual, and we wish that counterfactual stays above $50\%$ probability towards a desired class, then as per Section C of the \emph{supplementary material}, we have: 
\begin{align*}
 D_{\mathrm{KL}}(p_2(\omega \vert \cdot) \vert \vert p_1(\omega \vert \cdot)) \leq 0.10125, 
\end{align*}
is necessary for a counterfactual under a model change to confidently remain a counterfactual. In Figure \ref{fig:closedform} we verify equation \ref{ineq_closed2} showing the $D_{\mathrm{KL}}$ that is required for the theoretical bound to remain over 50\% probability of the desired class, where:
\begin{align}\label{ineq_closed2}
 D_{\mathrm{KL}}(p_2(\omega \vert \cdot) \vert \vert p_1(\omega \vert \cdot)) \lessapprox 0.111, 
\end{align}
as we substitute $\delta$ for $1 - p(y^\prime \vert \mathbf{x}^\prime, \mathcal{D}_{prev})$.  

It is worth noting, this is not limited to the applications of counterfactuals as any $\delta$ or $p(y^\prime \vert \mathbf{x}^\prime, \mathcal{D}_{prev})$ can be substituted. 

As clear from the proof of Theorem \ref{main_lemma_sup}, under an observed predictive posterior $p_1(y^\prime \vert \mathbf{x}^\prime, \mathcal{D}_{prev})$, we have the less conservative bound: 
\begin{align*}
  &p_2(y^\prime \vert \mathbf{x}^\prime, \mathcal{D}_{prev} \cup \mathcal{D}_{new}) \geq p_1(y^\prime \vert \mathbf{x}^\prime, \mathcal{D}_{prev}) - 2 \cdot \sqrt{\frac{1}{2} D_{\mathrm{KL}}(p_2(\omega \vert \cdot) \Vert p_1(\omega \vert \cdot))}.
\end{align*}

\begin{figure}[h]
    \centering
    \includegraphics[width=0.65\linewidth]{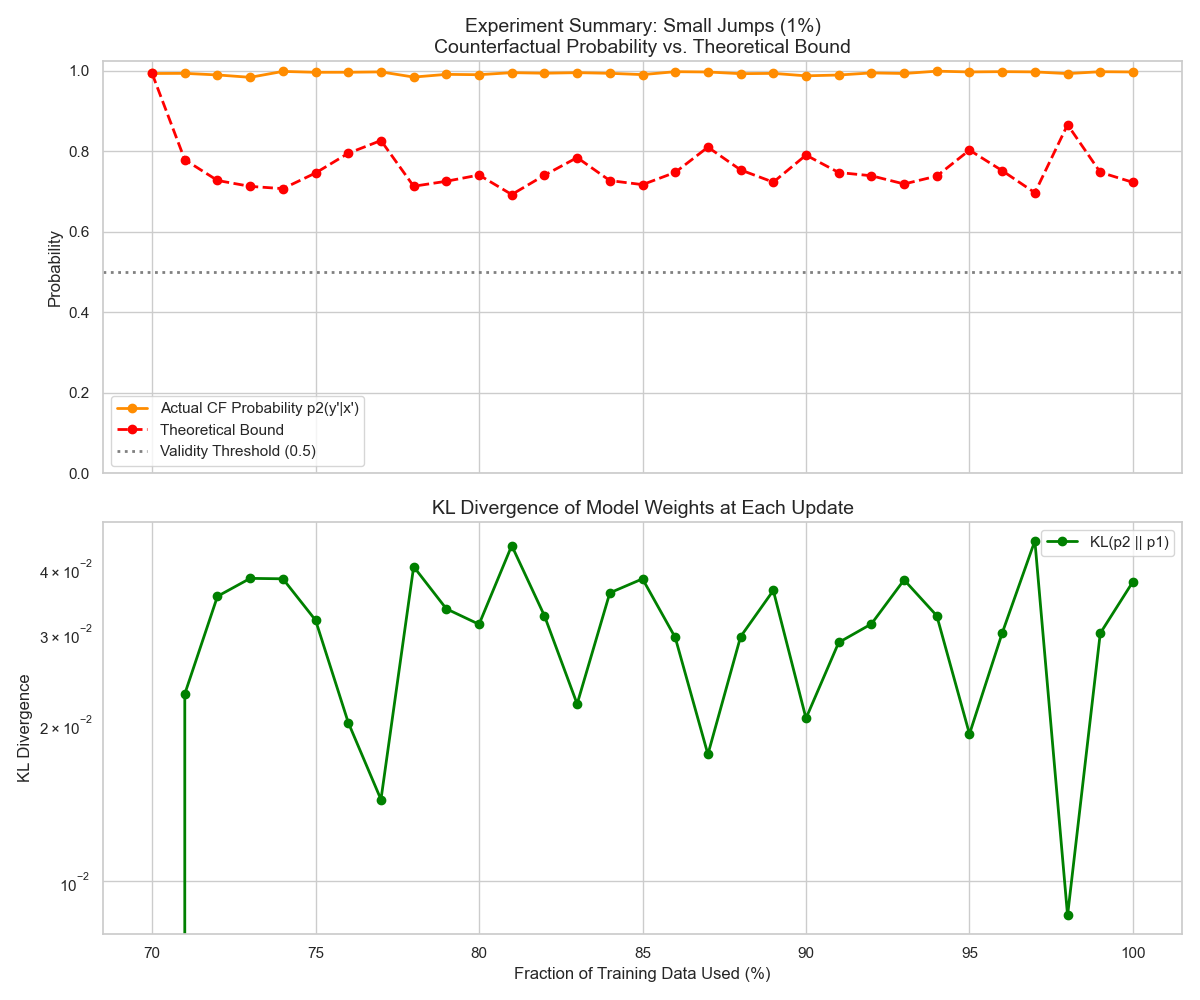}
    \caption{1\% increments in training data evaluating the theoretical bound vs computed approximate bound subject to the posterior over model weights.}
    \label{fig:1inc_e4}
\end{figure}
We empirically explore this in the results section and the \emph{supplementary material}. We can also ensure a conservative upper-bound on the predictive variance, leading to Theorem \ref{variance_upper_bound}. 

\begin{theorem}\label{variance_upper_bound}
Let $\text{Var}_{\omega \sim p_1(\omega\vert \mathcal{D}_{prev})}[p_1(y^\prime \vert \mathbf{x}^\prime, \omega)]$ be the variance in likelihood over a parametric posterior distribution $p_1(\omega\vert \mathcal{D}_{prev})$ given previously observed data $\mathcal{D}_{prev}$, where counterfactual $\mathbf{x}^\prime$ is $\epsilon$-robust, such that:
\begin{align*}
    \text{Var}_{\omega \sim p_1(\omega\vert \mathcal{D}_{prev})}[p_1(y^\prime \vert \mathbf{x}^\prime, \omega)] \leq \epsilon,
\end{align*}
then under new observed data $\mathcal{D}_{new}$, providing a new posterior distribution of model parameters $p_2(\omega \vert\mathcal{D}_{prev} \cup \mathcal{D}_{new})$, we have: 
\begin{align*}
\text{Var}_{\omega \sim p_2(\omega\vert \mathcal{D}_{prev} \cup \mathcal{D}_{new})}&[p_2(y^\prime \vert \mathbf{x}^\prime, \omega)] \leq  
\epsilon + 6 \cdot \sqrt{\frac{1}{2}D_{\mathrm{KL}}(p_2(\omega \vert \cdot) \Vert p_1(\omega \vert \cdot))}.
\end{align*}
\end{theorem}
Similarly, Theorem \ref{variance_upper_bound} shows that the variance degrades in a controlled way, bounded by $\epsilon$ plus a term dependent on KL divergence. By the proof of Theorem 2, we see that for an observed $ \text{Var}_{\omega \sim p_1(\omega\vert \mathcal{D}_{prev})}[p_1(y^\prime \vert \mathbf{x}^\prime, \omega)]$, we have:
\begin{align*}
    \text{Var}_{\omega \sim p_2(\omega\vert \mathcal{D}_{prev} \cup \mathcal{D}_{new})}&[p_2(y^\prime \vert \mathbf{x}^\prime, \omega)] \leq  \text{Var}_{\omega \sim p_1(\omega\vert \mathcal{D}_{prev})}[p_1(y^\prime \vert \mathbf{x}^\prime, \omega)] +  6 \cdot \sqrt{\frac{1}{2}D_{\mathrm{KL}}(p_2(\omega \vert \cdot)\Vert p_1(\omega \vert \cdot))}.
\end{align*}
All missing proofs are provided in the \emph{supplementary material}.

\section{Experimental Setup}

In the experimental setup, we explicitly consider Bayesian-based counterfactual explainers. Thus, we limit our scope on what we refer to as Schut \citep{Schut2021GeneratingIC} (using the ``greedy" approach in algorithm 1 of their paper) and BayesCF \citep{batten2025uncertaintyaware} due to their Bayesian nature. For each counterfactual method we adopt BNNs\footnote{we use the BNN implementation from torchbnn: \url{https://pypi.org/project/torchbnn/} \citep{lee2022graddiv}} and MC Dropout (see \emph{supplementary material}) for all methods. Furthermore, for a fair comparison we accept all counterfactuals generated by the methods including our own, and do not explicitly accept $\delta$-safe or $\epsilon$-robust counterfactuals only.

We present evaluation metrics, model architecture and hyperparameter details in the \emph{supplementary material}

\subsection{Configuration}

All experiments are conducted with a 3.50 GHz AMD Ryzen 5 5600 6-Core CPU and an NVIDIA GeForce RTX 4060 GPU. The implementation of each method was experimented using PyTorch. 

\subsection{Datasets}
The datasets used in this paper follow standard counterfactual explanation literature \citep{batten2025uncertaintyaware, Schut2021GeneratingIC, 10.5555/3709347.3743791}, where we adopt the MNIST, German Credit, Wisconsin Breast Cancer and Spambase datasets, we also include a MedMNIST \citep{medmnistv1, medmnistv2} dataset, namely the PneumoniaMNIST. Dataset preprocessing details and further experiments are provided in the \emph{supplementary material} with an example counterfactual explanation.

\section{Results}
\subsection{Robustness Under Model Changes}
\begin{figure}
    \centering
    \includegraphics[width=0.65\linewidth]{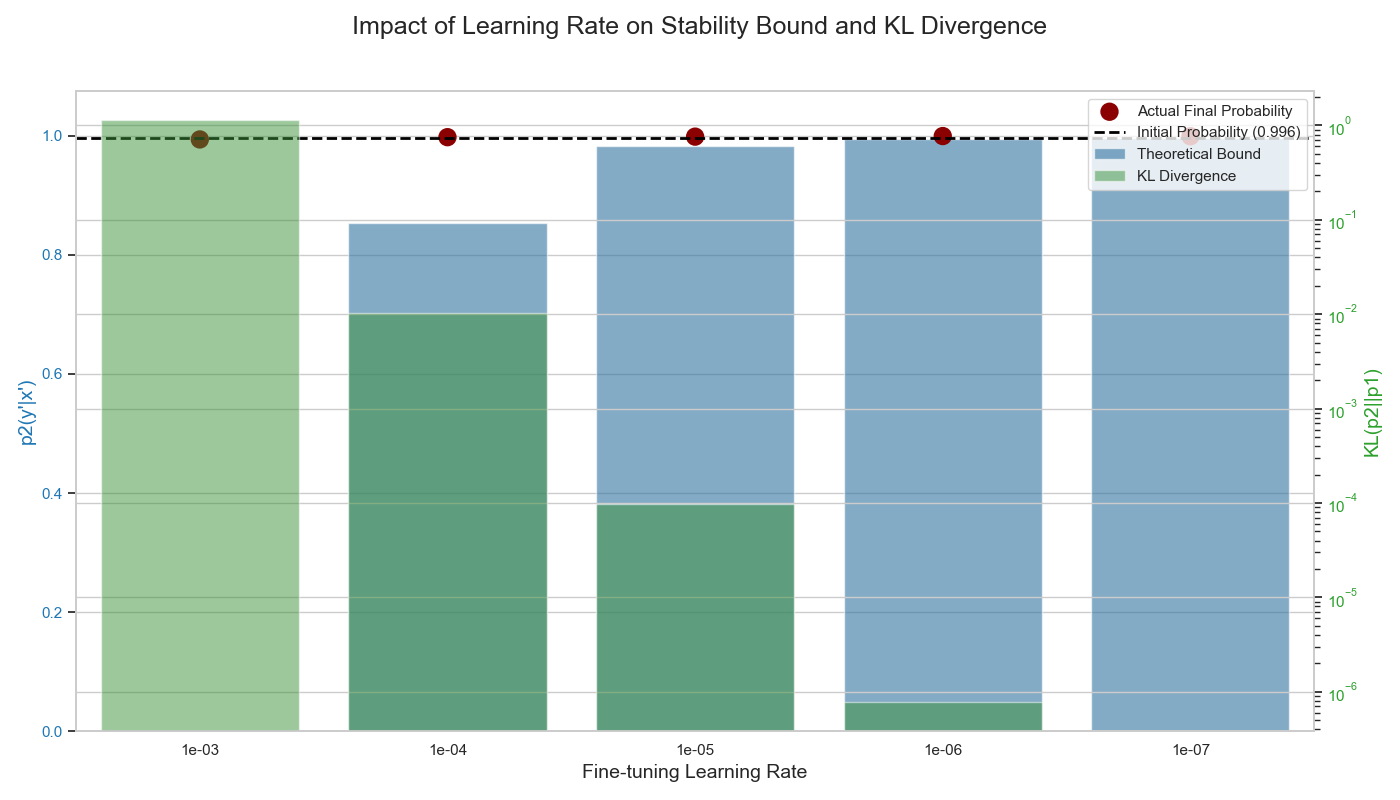}
    \caption{Comparison of the theoretical bound and the computed approximate bound under the posterior over model weights.}
    \label{fig:bounds_validity}
\end{figure}
To test the validity of Theorem 1, specifically the bound $
   p_1(y^\prime \vert \mathbf{x}^\prime, \mathcal{D}_{prev}) - 2 \cdot \sqrt{\frac{1}{2} D_{\mathrm{KL}}(p_2(\omega \vert \cdot) \Vert p_1(\omega \vert \cdot))}
$, we carried out controlled experiments with minimal parameter change. We first train a neural network on 95\% of the data to establish the initial posterior, $p_1(\omega \vert \cdot)$. Then, we incrementally update the model with an additional 1\% of the training data. For this fine-tuning, only the final layer is updated, using a small learning rate (1e-5) to produce the new posterior, $p_2(\omega \vert \cdot)$. The KL divergence, $D_{\mathrm{KL}}(p_2(\omega \vert \cdot) \Vert p_1(\omega \vert \cdot))$, is estimated by approximating both posteriors using a BNN for demonstration. We illustrate the relationship between the KL divergence, the theoretical bound, the counterfactual threshold, and the actual counterfactual probability under newly observed data in Figure \ref{fig:1inc_e4} and Table \ref{tab:incremental_results}. Additional experiments are presented in the \emph{supplementary material}, where we vary the learning rate and data increment size, and analyze their effect on the KL divergence and the theoretical bound.

In Figure \ref{fig:bounds_validity}, we observe that a learning rate of $10^{-5}$ yields a reasonably tight bound. This confirms that sufficiently small learning rates are required in practice to obtain meaningful bounds. The bound becomes tighter as model changes decrease in size. This observation motivated our choice of learning rate for model changes in this work.

In summary, a small learning rate is necessary during data updates to ensure that parameter changes remain small enough for the bound to be meaningful.

\begin{table*}[ht]
\centering
\caption{Performance metrics measured across all datasets. The best performing method is highlighed in \textbf{bold} text. The experiments are run over 5 runs and the \emph{mean} and \emph{standard deviation} are recorded. Each run is evaluated over 100 instances from the test data. These results use \textbf{BNN}.}
\label{tab:flipped_results}
\begin{tabular}{llccc}
\toprule
\textbf{Dataset} & \textbf{Metric} & \textbf{PSCE (Ours)} & \textbf{BayesCF} & \textbf{Schut}\\
\midrule

\multirow{4}{*}{Credit}
  & IM1 $(\downarrow)$              & \textbf{0.6561 $\pm$ 0.0603}  &  0.8311 $\pm$ 0.0289 &  0.9042 $\pm$ 0.0222 \\
  & Implausibility $(\downarrow)$   & \textbf{8.4606 $\pm$ 0.2683}  & 10.1530 $\pm$ 0.3799  & 10.8363 $\pm$ 0.1218  \\
  & Robustness Ratio $1e\text{-}3$ $(\downarrow)$ & \textbf{0.1400 $\pm$ 0.0203} & 0.1461 $\pm$ 0.0168  & 0.2318 $\pm$ 0.0304  \\
  & Validity \% $(\uparrow)$        & \textbf{100.0 $\pm$ 0.0} &  \textbf{100.0 $\pm$ 0.0} & \textbf{100.0 $\pm$ 0.0}  \\
\midrule

\multirow{4}{*}{Breast Cancer}
  & IM1 $(\downarrow)$              & 1.5540 $\pm$ 0.2427  &  1.1957 $\pm$ 0.0507 & \textbf{1.1151 $\pm$ 0.0362}  \\
  & Implausibility $(\downarrow)$   & \textbf{5.7579 $\pm$ 0.0439} &  7.2014 $\pm$ 0.0423 & 7.4072 $\pm$ 0.0333\\
  & Robustness Ratio $1e\text{-}3$ $(\downarrow)$ & \textbf{ 0.1566 $\pm$ 0.0152} & 0.3867 $\pm$ 0.0306  &  0.3050 $\pm$ 0.0211 \\
  & Validity \% $(\uparrow)$        & \textbf{86.8$\pm$3.2}  & 78.4$\pm$3.5  & 84.8$\pm$1.9  \\
\midrule
\multirow{4}{*}{MNIST}
  & IM1 $(\downarrow)$              & \textbf{0.9768 $\pm$ 0.1598}   &  1.6408 $\pm$ 0.1267 &  1.2298 $\pm$ 0.0526  \\
  & Implausibility $(\downarrow)$   &  \textbf{28.1567 $\pm$ 0.2665}& 35.7763 $\pm$ 0.0737  &  31.2247 $\pm$ 0.2067 \\
  & Robustness Ratio $1e\text{-}3$ $(\downarrow)$ & \textbf{0.7822 $\pm$ 0.0289} &  1.0191 $\pm$ 0.0463 &  0.9011 $\pm$ 0.0170 \\
  & Validity \% $(\uparrow)$        &  \textbf{98.2$\pm$1.6}&  67.6$\pm$5.4  &  96.0$\pm$1.8 \\
\midrule

\multirow{4}{*}{Spam}
  & IM1 $(\downarrow)$              & \textbf{0.7387 $\pm$ 0.0578}  &  1.0948 $\pm$ 0.0218 & 0.9137 $\pm$ 0.0299  \\
  & Implausibility $(\downarrow)$   &  \textbf{ 6.1781 $\pm$ 0.0312} & 9.4032 $\pm$ 0.0471  & 12.9145 $\pm$ 0.1061  \\
  & Robustness Ratio $1e\text{-}3$ $(\downarrow)$ & \textbf{0.1028 $\pm$ 0.0099}  & 0.1507 $\pm$ 0.0124  & 0.2046 $\pm$ 0.0083 \\
  & Validity \% $(\uparrow)$        &  96.0 $\pm$ 3.3 & \textbf{100.0 $\pm$ 0.0}  & 98.6$\pm$0.5  \\
  \bottomrule
  \multirow{4}{*}{PneumoniaMNIST}
  & IM1 $(\downarrow)$              & \textbf{0.630 $\pm$ 0.095} & 1.062 $\pm$ 0.009  & 1.157 $\pm$ 0.024   \\
  & Implausibility $(\downarrow)$   & \textbf{4.465 $\pm$ 0.332} &  5.687 $\pm$ 0.027 & 5.350 $\pm$ 0.019 \\
  & Robustness Ratio $1e\text{-}3$ $(\downarrow)$ & \textbf{0.255 $\pm$ 0.020}  &  1.191 $\pm$ 0.009  &  1.015 $\pm$ 0.033\\
  & Validity \% $(\uparrow)$        & \textbf{99.8$\pm$0.4}   & 98.4$\pm$1.5  & 91.2$\pm$3.6   \\
\bottomrule
\end{tabular}
\end{table*}

\begin{table*}[ht]
\centering
\caption{Performance metrics measured across all datasets. The best performing method is highlighed in \textbf{bold} text. The experiments are run over 5 runs and the \emph{mean} and \emph{standard deviation} are recorded. Each run is evaluated over 100 instances from the test data. These results use \textbf{MC Dropout}.}
\label{tab:flipped_results2}
\begin{tabular}{llccc}
\toprule
\textbf{Dataset} & \textbf{Metric} & \textbf{PSCE (Ours)} & \textbf{BayesCF} & \textbf{Schut} \\
\midrule

\multirow{4}{*}{Credit}
  & IM1 $(\downarrow)$              &  \textbf{0.7383 $\pm$ 0.0341} & 1.0767 $\pm$ 0.0214     & 0.9251 $\pm$ 0.0320  \\
  & Implausibility $(\downarrow)$   &  \textbf{8.6113 $\pm$ 0.1494}&  9.3615 $\pm$ 0.0543  & 9.9943 $\pm$ 0.2124  \\
  & Robustness Ratio $1e\text{-}3$ $(\downarrow)$ &  \textbf{0.0622 $\pm$ 0.0103}& 0.1219 $\pm$ 0.0106  &  0.1418 $\pm$ 0.0208\\
  & Validity \% $(\uparrow)$        & \textbf{100.0$\pm$0.0}  & 66.8$\pm$ 5.2  &  \textbf{100.0$\pm$0.0}  \\
\midrule

\multirow{4}{*}{Breast Cancer}
  & IM1 $(\downarrow)$              & \textbf{0.6932 $\pm$ 0.0911}  &  1.1300 $\pm$ 0.0306 &  1.0332 $\pm$ 0.0340 \\
  & Implausibility $(\downarrow)$   & \textbf{5.6275 $\pm$ 0.0627} &  7.1505 $\pm$ 0.0323 & 7.2913 $\pm$ 0.0354 \\
  & Robustness Ratio $1e\text{-}3$ $(\downarrow)$ & 0.0800 $\pm$ 0.0047  & \textbf{0.0539 $\pm$ 0.0104}  &  0.1045 $\pm$ 0.0105 \\
  & Validity \% $(\uparrow)$        &  \textbf{97.2$\pm$1.6} &  79.2$\pm$5.0 & 90.4$\pm$0.8 \\
\midrule

\multirow{4}{*}{MNIST}
  & IM1 $(\downarrow)$              &  \textbf{0.9273 $\pm$ 0.0172}  & 1.6426 $\pm$ 0.2774 & 1.2373 $\pm$ 0.0392  \\
  & Implausibility $(\downarrow)$   & \textbf{27.0337 $\pm$ 0.4335} & 32.9022 $\pm$ 0.2163  &  31.3665 $\pm$ 0.1397 \\
  & Robustness Ratio $1e\text{-}3$ $(\downarrow)$ & \textbf{0.6493 $\pm$ 0.0496}  &  0.9918 $\pm$ 0.0471  &  1.0696 $\pm$ 0.0412 \\
  & Validity \% $(\uparrow)$        &  99.6$\pm$0.8 & 47.8$\pm$3.3  & \textbf{100.0 $\pm$ 0.0}   \\
\midrule
\multirow{4}{*}{Spam}
  & IM1 $(\downarrow)$              &   \textbf{0.8967 $\pm$ 0.0316}  & 1.4798 $\pm$ 0.0122    &  1.2756 $\pm$ 0.0225  \\
  & Implausibility $(\downarrow)$   & \textbf{7.2057 $\pm$ 0.0618}  &   9.4627 $\pm$ 0.0442  &  10.0120 $\pm$ 0.0461 \\
  & Robustness Ratio $1e\text{-}3$ $(\downarrow)$ & \textbf{0.0429 $\pm$ 0.0063} &  0.0868 $\pm$ 0.0092   &   0.0870 $\pm$ 0.0151 \\
  & Validity \% $(\uparrow)$        &  91.8 $\pm$ 4.5 & 79.0$\pm$4.0  &  \textbf{99.0$\pm$0.0} \\
  \midrule
  \multirow{4}{*}{PneumoniaMNIST}
  & IM1 $(\downarrow)$              &  \textbf{0.545 $\pm$ 0.102}  &  1.062 $\pm$ 0.012 &  1.031 $\pm$ 0.007  \\
  & Implausibility $(\downarrow)$   & \textbf{4.489 $\pm$ 0.149}  & 5.647 $\pm$ 0.037  & 6.066 $\pm$ 0.138 \\
  & Robustness Ratio $1e\text{-}3$ $(\downarrow)$ &  \textbf{0.098 $\pm$ 0.017} &  0.597 $\pm$ 0.030 & 0.770 $\pm$ 0.034 \\
  & Validity \% $(\uparrow)$        & 99.4$\pm$0.5 & \textbf{100.0$\pm$0.0} & \textbf{100.0$\pm$0.0}   \\
\bottomrule
\end{tabular}
\end{table*}

\subsection{Comparison Under Key Metrics}
In Table \ref{tab:flipped_results} and Table \ref{tab:flipped_results2}, we observe the performance of the proposed PSCE when compared against existing Bayesian CEs. Here we can see our approach mostly exhibits superior performance across key metrics.

\section{Conclusion}

This paper introduces the PSCE method for Bayesian inspired CEs. We show both theoretically and empirically, the robustness of CEs under our method with respect to model changes. Our approach empirically performs better than existing Bayesian inspired counterfactual explanation methods on a collection of core datasets that are used within counterfactual literature.

\bibliographystyle{apalike}
\bibliography{references}

\newpage
\appendix 
\section{MISSING PROOFS}

\subsection{Proof of Theorem 1}
Recalling the Theorem for readability:
Let $p_1(\omega \vert \mathcal{D}_{\text{prev}})$ be a probabilistic model's posterior distribution over its parameters $\omega$. After a data update, the new posterior is $p_2(\omega \vert \mathcal{D}_{\text{prev}} \cup \mathcal{D}_{\text{new}})$. Given the initial posterior predictive probability for a class $y^\prime$ given a counterfactual $\mathbf{x}^\prime$ is at least $p_1(y^\prime \vert \mathbf{x}^\prime, \mathcal{D}_{prev}) \geq 1 - \delta$, then the updated posterior predictive probability is bounded such that:
\begin{align*}
p_2(y^\prime \vert \mathbf{x}^\prime, \mathcal{D}_{prev} \cup \mathcal{D}_{new}) \geq (1 - \delta) - 2\cdot \sqrt{\frac{1}{2}D_{\mathrm{KL}}(p_2(\omega \vert \cdot) \vert \vert p_1(\omega \vert \cdot))},
\end{align*}
where, $p_2(\omega \vert \cdot) = p_2(\omega \vert \mathcal{D}_{prev} \cup \mathcal{D}_{new})$ and $p_1(\omega \vert \cdot) = p_1(\omega \vert \mathcal{D}_{prev})$.

\begin{proof}
Let $\pi_2 = p_2(y^\prime \vert \mathbf{x}^\prime, \mathcal{D}_{prev} \cup \mathcal{D}_{new})$ and $\pi_1 = p_1(y^\prime \vert \mathbf{x}^\prime, \mathcal{D}_{prev})$. We also let $f(\omega) = p(y^\prime \vert \mathbf{x}^\prime, \omega)$. Then,

\begin{align*}
 \vert \pi_2 - \pi_1\vert &= \vert \mathbb{E}_{\omega \sim p_2(\omega \vert \cdot)}[f(\omega)] - \mathbb{E}_{\omega \sim p_1(\omega \vert \cdot)}[f(\omega)] \vert \\
\\ &= \left\vert \int f(\omega)p_2(\omega \vert \cdot)d\omega - \int f(\omega)p_1(\omega \vert \cdot)d\omega \right\vert
\\
    &= \left\vert \int f(\omega) \big(p_2(\omega \vert \cdot) - p_1(\omega \vert \cdot)\big) d\omega \right\vert.
\end{align*}
By properties of the integral we can note that:
\begin{align*}
&\left\vert \int f(\omega) \big(p_2(\omega \vert \cdot)- p_1(\omega \vert \cdot)\big) d\omega \right\vert \ \leq \sup_{\omega} \vert f(\omega)\vert \cdot \int \vert p_2(\omega \vert \cdot) - p_1(\omega \vert \cdot)\vert d\omega
\end{align*}
then by the total variation (TV), namely: $\text{TV}(p_2(\omega \vert \cdot), p_1(\omega \vert \cdot)) = \frac{1}{2} \int \vert p_2(\omega \vert \cdot) - p_1(\omega \vert \cdot) \vert$, we have:
\begin{align*}
\vert\pi_2 - \pi_1\vert &\leq
2\cdot \sup_{\omega} \vert f(\omega)\vert \cdot \text{TV}(p_2(\omega \vert \cdot), p_1(\omega \vert \cdot)), 
\end{align*}
Then by Pinsker's inequality, we have:
\begin{align*}
\text{TV}(p_2(\omega \vert \cdot), p_1(\omega \vert \cdot)) \leq \sqrt{\frac{1}{2} D_{\mathrm{KL}}(p_2(\omega \vert \cdot) \vert \vert p_1(\omega \vert \cdot))},
\end{align*}
thus,
\begin{align*}
\vert \pi_2 - \pi_1\vert \leq 2\cdot\sup_{\omega} \vert f(\omega)\vert \cdot \sqrt{\frac{1}{2} D_{\mathrm{KL}}(p_2(\omega \vert \cdot) \Vert p_1(\omega \vert \cdot))}.
\end{align*}
Since $\pi_1 \in [1-\delta, 1] (\text{implying that }\pi_1 \geq 1-\delta)$ for some $\delta \in [0,1]$, that is all that is necessary to establish the lower bound:
\begin{align*}
\pi_2 &\geq \pi_1 - 2\cdot\sup_{\omega} \ p(y^\prime \vert \mathbf{x}^\prime, \omega) \cdot \sqrt{\frac{1}{2} D_{\mathrm{KL}}(p_2(\omega \vert \cdot) \Vert p_1(\omega \vert \cdot))} \\ &\geq (1-\delta) - 2\cdot \sup_{\omega} \ p(y^\prime \vert \mathbf{x}^\prime, \omega) \cdot \sqrt{\frac{1}{2} D_{\mathrm{KL}}(p_2(\omega \vert \cdot) \Vert p_1(\omega \vert \cdot))},
\end{align*}
finally, since we know that $p(y^\prime \vert \mathbf{x}^\prime, \omega) \in [0,1]$, then:
\begin{align*}
  (1-\delta) - 2 \cdot \sqrt{\frac{1}{2} D_{\mathrm{KL}}(p_2(\omega \vert \cdot) \Vert p_1(\omega \vert \cdot))},
\end{align*}
providing the final bound.
\end{proof}

\subsection{Proof of Theorem 2}
We begin by again recalling Theorem 2 for readability: Let $\text{Var}_{\omega \sim p_1(\omega\vert \mathcal{D}_{prev})}[p_1(y^\prime \vert \mathbf{x}^\prime, \omega)]$ be the variance in likelihood over a parametric distribution $p_1(\omega\vert \mathcal{D}_{prev})$ given previously observed data $\mathcal{D}_{prev}$, where counterfactual $\mathbf{x}^\prime$ is $\epsilon$-robust, such that:
\begin{align*}
    \text{Var}_{\omega \sim p_1(\omega\vert \mathcal{D}_{prev})}[p_1(y^\prime \vert \mathbf{x}^\prime, \omega)] \leq \epsilon,
\end{align*}
then under new observed data $\mathcal{D}_{new}$, providing a new posterior distribution of model parameters $p_2(\omega \vert\mathcal{D}_{prev} \cup \mathcal{D}_{new})$, we have: 
\begin{align*}
\text{Var}_{\omega \sim p_2(\omega\vert \mathcal{D}_{prev} \cup \mathcal{D}_{new})}&[p_2(y^\prime \vert \mathbf{x}^\prime, \omega)] \leq  \epsilon + 6 \cdot \sqrt{\frac{1}{2}D_{\mathrm{KL}}(p_2(\omega \vert \cdot) \Vert p_1(\omega \vert \cdot))}.
\end{align*}
\begin{proof}
 By adopting the previous notation from the proof of Theorem 1 for simplicity, such that $f(\omega) := p(y^\prime \vert \mathbf{x}^\prime, \omega)$. We can observe that: 
\begin{align*}
\text{Var}_{\omega \sim p_1(\omega\vert \mathcal{D}_{prev})}[f(\omega)] &= \mathbb{E}[(f(\omega)  - \mathbb{E}[f(\omega)])^2] \\ 
&= \mathbb{E}[f(\omega)^2 - 2f(\omega)\mathbb{E}[f(\omega)] + \mathbb{E}[f(\omega)]^2 ] \\
&= \mathbb{E}[f(\omega)^2] - 2\mathbb{E}[f(\omega)]\mathbb{E}[f(\omega)] + \mathbb{E}[f(\omega)]^2 \\ 
&= \mathbb{E}[f(\omega)^2] - 2\mathbb{E}[f(\omega)]^2 + \mathbb{E}[f(\omega)]^2 \\ 
&= \mathbb{E}[f(\omega)^2] - \mathbb{E}[f(\omega)]^2 : \mathbb{E} := \mathbb{E}_{\omega \sim p_1(\omega\vert \mathcal{D}_{prev})}
\end{align*} 
Similarly, 
\begin{align*}
\text{Var}_{\omega \sim p_2(\omega\vert \mathcal{D}_{prev} \cup \mathcal{D}_{new})}[f(\omega)] = \mathbb{E}[f(\omega)^2] - \mathbb{E}[f(\omega)]^2 : \mathbb{E} := \mathbb{E}_{\omega \sim p_2(\omega\vert \mathcal{D}_{prev} \cup \mathcal{D}_{new})}
\end{align*} 
Then, similar to Theorem 1, we aim to find a bound on $\vert \text{Var}_{\omega \sim p_2(\omega \vert \cdot)}[f(\omega)] - \text{Var}_{\omega \sim p_1(\omega \vert \cdot)}[f(\omega)] \vert$, where  $p_2(\omega \vert \cdot) = p_2(\omega \vert \mathcal{D}_{prev} \cup \mathcal{D}_{new})$ and $p_1(\omega \vert \cdot) = p_1(\omega \vert \mathcal{D}_{prev})$, such that:
\begin{align*}
    &\vert \text{Var}_{\omega \sim p_2(\omega \vert \cdot)}[f(\omega)] - \text{Var}_{\omega \sim p_1(\omega \vert \cdot)}[f(\omega)] \vert =  \\
    &\vert (\mathbb{E}_{\omega \sim p_2(\omega \vert \cdot)}[f(\omega)^2] - \mathbb{E}_{\omega \sim p_2(\omega \vert \cdot)}[f(\omega)]^2) - (\mathbb{E}_{\omega \sim p_1(\omega \vert \cdot)}[f(\omega)^2] - \mathbb{E}_{\omega \sim p_1(\omega \vert \cdot)}[f(\omega)]^2)\vert \\ &\leq
    \vert \mathbb{E}_{\omega \sim p_2(\omega \vert \cdot)}[f(\omega)^2] - \mathbb{E}_{\omega \sim p_1(\omega \vert \cdot)}[f(\omega)^2] \vert + \vert \mathbb{E}_{\omega \sim p_2(\omega \vert \cdot)}[f(\omega)]^2 - \mathbb{E}_{\omega \sim p_1(\omega \vert \cdot)}[f(\omega)]^2  \vert
\end{align*}
by the triangle inequality. Then observing $f(\omega) \in [0,1]$, this implies that $f(\omega)^2 \in [0,1]$, therefore, by total variation (TV) as per Theorem 1, we similarly observe that: 
\begin{align*}
     \vert \mathbb{E}_{\omega \sim p_2(\omega \vert \cdot)}[f(\omega)^2] - \mathbb{E}_{\omega \sim p_1(\omega \vert \cdot)}[f(\omega)^2] \vert \leq 2 \cdot \sqrt{\frac{1}{2}D_{\mathrm{KL}}(p_2 \Vert p_1(\omega \vert \cdot))},
\end{align*}
we can then proceed to consider the term  \begin{align*}
&\vert \mathbb{E}_{\omega \sim p_2(\omega \vert \cdot)}[f(\omega)]^2 - \mathbb{E}_{\omega \sim p_1(\omega \vert \cdot)}[f(\omega)]^2  \vert = \vert \mathbb{E}_{\omega \sim p_2(\omega \vert \cdot)}[f(\omega)] - \mathbb{E}_{\omega \sim p_1(\omega \vert \cdot)}[f(\omega)]  \vert \vert \mathbb{E}_{\omega \sim p_2(\omega \vert \cdot)}[f(\omega)] + \mathbb{E}_{\omega \sim p_1(\omega \vert \cdot)}[f(\omega)]  \vert 
\end{align*}
then since, we know that both, $0 \leq \mathbb{E}_{\omega \sim p_2(\omega \vert \cdot)}[f(\omega)] \leq 1$ and $0 \leq \mathbb{E}_{\omega \sim p_1(\omega \vert \cdot)}[f(\omega)] \leq 1$, we can see that: 
\begin{align*}
&\vert \mathbb{E}_{\omega \sim p_2(\omega \vert \cdot)}[f(\omega)] - \mathbb{E}_{\omega \sim p_1(\omega \vert \cdot)}[f(\omega)]  \vert \underbrace{\vert \mathbb{E}_{\omega \sim p_2(\omega \vert \cdot)}[f(\omega)] + \mathbb{E}_{\omega \sim p_1(\omega \vert \cdot)}[f(\omega)]}_{\leq 2}  \vert     
\implies  \vert \mathbb{E}_{\omega \sim p_2(\omega \vert \cdot)}[f(\omega)]^2 - \mathbb{E}_{\omega \sim p_1(\omega \vert \cdot)}[f(\omega)]^2  \vert \\ &\leq 2 \cdot \vert \mathbb{E}_{\omega \sim p_2(\omega \vert \cdot)}[f(\omega)] - \mathbb{E}_{\omega \sim p_1(\omega \vert \cdot)}[f(\omega)]  \vert \leq 4 \cdot \sqrt{\frac{1}{2}D_{\mathrm{KL}}(p_2(\omega \vert \cdot)\Vert p_1(\omega \vert \cdot))}
\\
& \implies \vert \mathbb{E}_{\omega \sim p_2(\omega \vert \cdot)}[f(\omega)]^2 - \mathbb{E}_{\omega \sim p_1(\omega \vert \cdot)}[f(\omega)]^2  \vert \leq 4 \cdot \sqrt{\frac{1}{2}D_{\mathrm{KL}}(p_2(\omega \vert \cdot)\Vert p_1(\omega \vert \cdot))} 
\end{align*} 
then by summing over the constants for the individual bounds, we have the final bound:
\begin{align*}
    \vert \text{Var}_{\omega \sim p_2(\omega \vert \cdot)}[f(\omega)] - \text{Var}_{\omega \sim p_1(\omega \vert \cdot)}[f(\omega)] \vert \leq  6 \cdot \sqrt{\frac{1}{2}D_{\mathrm{KL}}(p_2\Vert p_1(\omega \vert \cdot))} 
\end{align*}
and therefore: 
\begin{align*}
    \text{Var}_{\omega \sim p_2(\omega \vert \cdot)}[f(\omega)] &\leq \text{Var}_{\omega \sim p_1(\omega \vert \cdot)}[f(\omega)]  +  6 \cdot \sqrt{\frac{1}{2}D_{\mathrm{KL}}(p_2(\omega \vert \cdot) \Vert p_ 1(\omega))} \\ &\implies     \text{Var}_{\omega \sim p_2(\omega \vert \cdot)}[f(\omega)] \leq \epsilon  +  6 \cdot \sqrt{\frac{1}{2}D_{\mathrm{KL}}(p_2\Vert p_1(\omega \vert \cdot))} 
\end{align*}
\end{proof}

\section{ELBO Derivation}
We provide the standard ELBO derivation for completeness. Recall, we have that:
\begin{align*}
   \text{log }p(\mathbf{x}) &= \text{log }\int p_{\theta}(\mathbf{x}\vert\mathbf{z})p(\mathbf{z})\\ &= \text{log }\int q_{\phi}(\mathbf{z} \vert \mathbf{x})\frac{p_{\theta}(\mathbf{x}\vert\mathbf{z})p(\mathbf{z})}{q_{\phi}(\mathbf{z} \vert \mathbf{x})} d \mathbf{z} \\ 
   &\geq \int q_{\phi}(\mathbf{z} \vert \mathbf{x}) \text{log} \bigg( \frac{p_{\theta} (\mathbf{x}\vert\mathbf{z})p(\mathbf{z})}{q_{\phi}(\mathbf{z} \vert \mathbf{x})} \bigg) d \mathbf{z} \\ 
    &= \mathbb{E}_{q_{\phi}(\mathbf{z} \vert \mathbf{x})}\bigg[\text{log} \bigg( \frac{p_{\theta} (\mathbf{x}\vert\mathbf{z})p(\mathbf{z})}{q_{\phi}(\mathbf{z} \vert \mathbf{x})} \bigg)\bigg]
   \\
   &= \mathbb{E}_{q_{\phi}(\mathbf{z} \vert \mathbf{x})}[\text{log }p_{\theta}(\mathbf{x} \vert \mathbf{z})] + \mathbb{E}_{q_{\phi}(\mathbf{z} \vert \mathbf{x})}[\text{log }p(\mathbf{z})] - \mathbb{E}_{q_{\phi}(\mathbf{z} \vert \mathbf{x})}[\text{log } q_{\phi}(\mathbf{z} \vert \mathbf{x})] \\
   &= \mathbb{E}_{q_{\phi}(\mathbf{z} \vert \mathbf{x})}[\text{log }p_{\theta}(\mathbf{x} \vert \mathbf{z})] + \bigg( \int  q_{\phi}(\mathbf{z} \vert \mathbf{x}) \text{log } \bigg( \frac{p(\mathbf{z})}{q_{\phi}(\mathbf{z} \vert \mathbf{x})}
   \bigg) d\mathbf{z} \bigg) \\
   &= \mathbb{E}_{q_{\phi}(\mathbf{z} \vert \mathbf{x})}[\text{log }p_{\theta}(\mathbf{x} \vert \mathbf{z})] + \bigg( - D_{\mathrm{KL}}(q_{\phi}(\mathbf{z} \vert \mathbf{x}) \Vert p(\mathbf{z})) \bigg) \\ 
   &= \mathbb{E}_{q_{\phi}(\mathbf{z} \vert \mathbf{x})}[\text{log }p_{\theta}(\mathbf{x} \vert \mathbf{z})] - D_{\mathrm{KL}}(q_{\phi}(\mathbf{z} \vert \mathbf{x}) \Vert p(\mathbf{z})) \\ 
   &= \text{ELBO} \\ 
   &\implies \text{log }p(\mathbf{x}) \geq \text{ELBO} = \mathbb{E}_{q_{\phi}(\mathbf{z} \vert \mathbf{x})}[\text{log }p_{\theta}(\mathbf{x} \vert \mathbf{z})] - D_{\mathrm{KL}}(q_{\phi}(\mathbf{z} \vert \mathbf{x}) \Vert p(\mathbf{z})).
\end{align*} 
Completing the derivation. In the main paper, we use line 5 as our representation of the loss function and indeed follows the standard definition of ELBO. 

\section{Robustness of Bounds in Practice}
\begin{figure}
    \centering
    \includegraphics[width=0.45\linewidth]{images/bnn_images/summary_Small_Jumps_1.png}
    \includegraphics[width=0.45\linewidth]{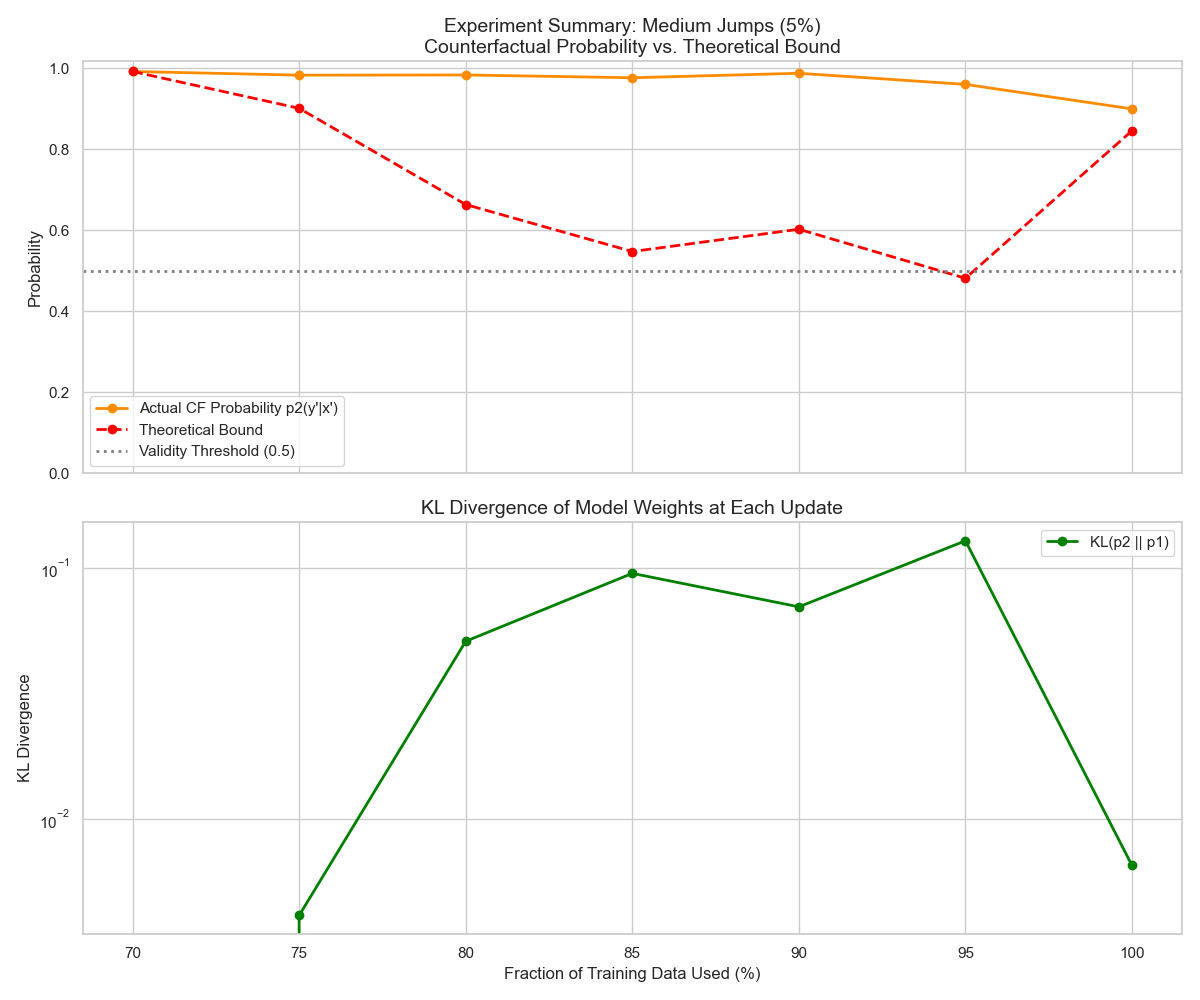}
    \caption{1\% and 5\% increments in training data.}
    \label{fig:1inc}
\end{figure}

In the main manuscript, we provide a simple analysis of the robustness under a small model change to the final layer of a network under new data. In Figures \ref{fig:1inc}, we illustrate the theoretical bound against the actual prediction probability under different learning different size data increments. 

 To extrapolate on the main manuscript, we work through the algebra on the RHS of the bound in Theorem 1. In practice, we wish to have a rational bounds above $\geq 0.5$ for a valid counterfactual, thus we can solve for this explicitly: 
\begin{align*}
    (1 - \delta) - 2\cdot \sqrt{\frac{1}{2}D_{\mathrm{KL}}(p_2(\omega \vert \cdot) \vert \vert p_1(\omega \vert \cdot))} \geq 0.5
\end{align*}
Then, working through the algebra to isolate the $D_{\mathrm{KL}}$ term, we have: 
\begin{align*}
    &(1 - \delta) - 2\cdot \sqrt{\frac{1}{2}D_{\mathrm{KL}}(p_2(\omega \vert \cdot) \vert \vert p_1(\omega \vert \cdot))} \geq 0.5 \\ 
    &\implies (1 - \delta) - 2\cdot \sqrt{\frac{1}{2}D_{\mathrm{KL}}(p_2(\omega \vert \cdot) \vert \vert p_1(\omega \vert \cdot))} - (1 - \delta) \geq 0.5 - (1 - \delta) \\ 
    &\implies  - 2\cdot \sqrt{\frac{1}{2}D_{\mathrm{KL}}(p_2(\omega \vert \cdot) \vert \vert p_1(\omega \vert \cdot))}  \geq \delta - 0.5 \\ 
    &\text{multiplying both sides by -1, we have:}
    \\ 
    &(-1)\bigg(- 2\cdot \sqrt{\frac{1}{2}D_{\mathrm{KL}}(p_2(\omega \vert \cdot) \vert \vert p_1(\omega \vert \cdot))}\bigg)  \leq (-1)(\delta - 0.5) \\ 
    &\implies 2\cdot \sqrt{\frac{1}{2}D_{\mathrm{KL}}(p_2(\omega \vert \cdot) \vert \vert p_1(\omega \vert \cdot))}  \leq 0.5 - \delta \\ 
    &\implies \frac{2\cdot \sqrt{\frac{1}{2}D_{\mathrm{KL}}(p_2(\omega \vert \cdot) \vert \vert p_1(\omega \vert \cdot))}}{2}  \leq \frac{(0.5 - \delta)}{2} \\ 
&\implies \sqrt{\frac{1}{2}D_{\mathrm{KL}}(p_2(\omega \vert \cdot) \vert \vert p_1(\omega \vert \cdot))}  \leq \frac{(0.5 - \delta)}{2} \\ 
&\implies \bigg(\sqrt{\frac{1}{2}D_{\mathrm{KL}}(p_2(\omega \vert \cdot) \vert \vert p_1(\omega \vert \cdot))}\bigg)^2  \leq \bigg(\frac{(0.5 - \delta)}{2}\bigg)^2 \\ 
&\implies \frac{1}{2}D_{\mathrm{KL}}(p_2(\omega \vert \cdot) \vert \vert p_1(\omega \vert \cdot)) \leq \bigg(\frac{(0.5 - \delta)}{2}\bigg)^2 \\ 
&\implies  2\cdot \bigg(\frac{1}{2}D_{\mathrm{KL}}(p_2(\omega \vert \cdot) \vert \vert p_1(\omega \vert \cdot))\bigg) \leq 2\cdot\bigg(\frac{(0.5 - \delta)}{2}\bigg)^2 \\ 
&\implies D_{\mathrm{KL}}(p_2(\omega \vert \cdot) \vert \vert p_1(\omega \vert \cdot)) \leq 2\cdot\bigg(\frac{(0.5 - \delta)}{2}\bigg)^2  \\ 
&\implies  D_{\mathrm{KL}}(p_2(\omega \vert \cdot) \vert \vert p_1(\omega \vert \cdot)) \leq \frac{(0.5 - \delta)^2}{2}
\end{align*}
Thus providing an ideal value for $D_{\mathrm{KL}}(p_2(\omega \vert \cdot) \vert \vert p_1(\omega \vert \cdot))$, if one wishes to ensure a counterfactual is equal to or over the probabilistic decision threshold for a counterfactual instance (0.5) upon a model change. Plugging in a value for $\delta = 0.05$, we can see for example, a counterfactual instance under model changes will remain a counterfactual instance, such that: 
\begin{align*}
    &D_{\mathrm{KL}}(p_2(\omega \vert \cdot) \vert \vert p_1(\omega \vert \cdot)) \leq \frac{(0.5 - 0.05)^2}{2} \\
    &\implies D_{\mathrm{KL}}(p_2(\omega \vert \cdot) \vert \vert p_1(\omega \vert \cdot)) \leq 0.10125 
\end{align*}

Thus, in the context of the $\delta$ used in this work (0.05) (one could also use $p_1(y^\prime \vert \mathbf{x}^\prime, \mathcal{D}_{prev})$ in place of $\delta$), the KL divergence between parameters of a model change must be $\leq 0.10125$ to have a meaningful bound in the context of counterfactuals. In practice, $p_2(\omega \vert \cdot)$ and $p_1(\omega \vert \cdot)$ are approximated. 

\section{Robustness of Variance to Model Changes}
We can go through a similar exercise to explore $\epsilon + 6 \cdot \sqrt{\frac{1}{2}D_{\mathrm{KL}}(p_2(\omega \vert \cdot) \Vert p_1(\omega \vert \cdot))}$, that is, consider we wish to have: 
\begin{align*}
    \epsilon + 6 \cdot \sqrt{\frac{1}{2}D_{\mathrm{KL}}(p_2(\omega \vert \cdot) \Vert p_1(\omega \vert \cdot))} \leq 0.01
\end{align*}
Then, again by isolating the $D_{\mathrm{KL}}$ term, we go through the algebra such that:
\begin{align*}
    &\epsilon + 6 \sqrt{\frac{1}{2} D_{\mathrm{KL}}(p_2(\omega \vert \cdot) \Vert p_1(\omega \vert \cdot))} \leq 0.01 \\ 
    &\implies 6 \sqrt{\frac{1}{2} D_{\mathrm{KL}}(p_2(\omega \vert \cdot) \Vert p_1(\omega \vert \cdot))} \leq 0.01 - \epsilon \\ 
    &\implies -6 \sqrt{\frac{1}{2} D_{\mathrm{KL}}(p_2(\omega \vert \cdot) \Vert p_1(\omega \vert \cdot))} \geq \epsilon - 0.01 \\ 
    &\implies \sqrt{\frac{1}{2} D_{\mathrm{KL}}(p_2(\omega \vert \cdot) \Vert p_1(\omega \vert \cdot))} \leq \frac{\epsilon - 0.01}{-6}
    \\ 
    &\implies \bigg(\sqrt{\frac{1}{2} D_{\mathrm{KL}}(p_2(\omega \vert \cdot) \Vert p_1(\omega \vert \cdot))}\bigg)^2  \leq \bigg(\frac{\epsilon - 0.01}{-6}\bigg)^2
    \\ &\implies \frac{1}{2} D_{\mathrm{KL}}(p_2(\omega \vert \cdot) \Vert p_1(\omega \vert \cdot)) \leq \bigg(\frac{\epsilon - 0.01}{-6}\bigg)^2 \\ 
    &\implies 2 \cdot \bigg(\frac{1}{2} D_{\mathrm{KL}}(p_2(\omega \vert \cdot) \Vert p_1(\omega \vert \cdot))\bigg) \leq 2 \cdot \bigg(\frac{\epsilon - 0.01}{-6}\bigg)^2 \\ 
    &\implies  D_{\mathrm{KL}}(p_2(\omega \vert \cdot) \Vert p_1(\omega \vert \cdot)) \leq \frac{2(\epsilon - 0.01)^2}{36} 
    = \frac{(\epsilon - 0.01)^2}{18}
\end{align*}
We do not proceed to provide more examples, but a similar exercise to Appendix C can be performed.

\section{Estimating KL Divergence Between BNN Posteriors}
\label{sec:kl_estimation}

Our robustness bounds (Theorems 1 and 2) require computing the KL divergence between
the posterior weight distributions before and after a model change. 
Since we use BNNs with mean-field Gaussian  approximations (via \texttt{torchbnn}), each parameter is modeled independently as
\begin{align*}
p(\omega_i) = \mathcal{N}(\mu_i, \sigma_i^2),
\end{align*}
where $\mu_i$ and $\sigma_i^2$ are learned by the BNN. We extract the means and variances from all \texttt{BayesLinear} layers using:

\begin{verbatim}
def get_bnn_posterior_params(model):
    mus, variances = [], []
    with torch.no_grad():
        for module in model.modules():
            if isinstance(module, bnn.BayesLinear):
                mus.append(module.weight_mu.flatten())
                variances.append(torch.exp(module.weight_log_sigma * 2).flatten())
                if module.bias_mu is not None:
                    mus.append(module.bias_mu.flatten())
                    variances.append(torch.exp(module.bias_log_sigma * 2).flatten())
    return torch.cat(mus), torch.cat(variances)
\end{verbatim}

This yields flattened tensors of all posterior means and variances across the network.

Given two BNNs with posterior parameters 
$(\mu_1, \sigma_1^2)$ and $(\mu_2, \sigma_2^2)$, the KL divergence is computed 
in closed form under the factorized Gaussian assumption:
\[
D_{\mathrm{KL}}(p_2 \vert\vert p_1)
= \sum_{i=1}^d 
\Bigg[
\log \frac{\sigma_{1,i}}{\sigma_{2,i}}
+ \frac{\sigma_{2,i}^2 + (\mu_{2,i} - \mu_{1,i})^2}{2\sigma_{1,i}^2}
- \frac{1}{2}
\Bigg].
\]

We implement the closed-form expression in terms of variances instead of standard deviation with the formula: 
\begin{verbatim}
def gaussian_diagonal_kl(mu_q, var_q, mu_p, var_p):
    return 0.5 * torch.sum(
        torch.log(var_p / var_q)
        + (var_q + (mu_q - mu_p).pow(2)) / var_p - 1)
\end{verbatim}

We compute $D_{\mathrm{KL}}(p_2(\omega \vert \cdot) \vert\vert p_1(\omega \vert \cdot))$ after each model change, 
where $p_1(\omega \vert \cdot)$ is the previous model posterior and $p_2(\omega \vert \cdot)$ is the updated posterior.
This value is then used in the bounds of Theorems 1 and 2 to assess the robustness 
of counterfactuals. In practice, this is an approximation as we use BNN's in our robustness experiments to approximate the true posterior parametric distributions, that is our $d$ model weights have parameters for the mean and variance learned approximate parametric posteriors. We note that, experiments in this paper regarding KL divergence between posteriors is calculated only for BNNs.  

\section{ADDITIONAL EXPERIMENTS}
\subsection{Supplementary $\delta$-Safe Example}
\begin{figure}[h]
\centering
\includegraphics[width=0.5\linewidth]{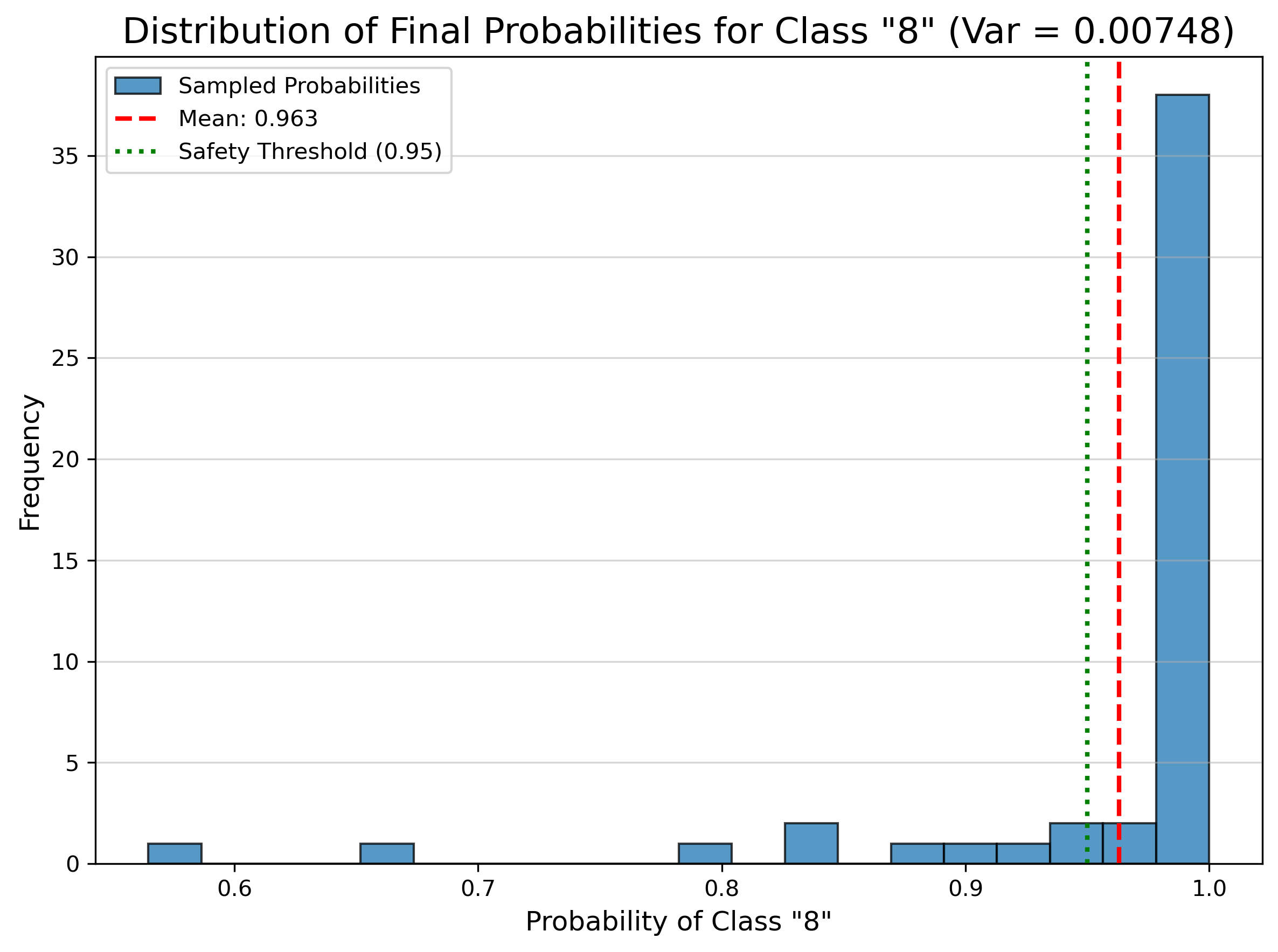}
\caption{$\delta$-safe and $\epsilon$-robust counterfactual produced by PSCE ($\delta = 0.05$, $\epsilon = 0.01$).}
\label{fig:distribution}
\end{figure}

\begin{figure*}
\centering
\includegraphics[width=0.95\linewidth]{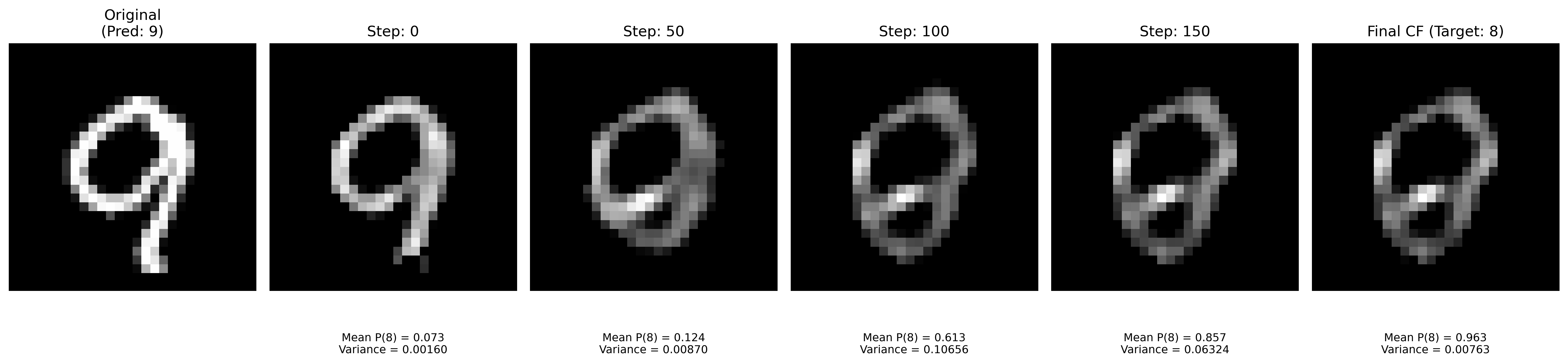}
\caption{Counterfactual generation process on the MNIST dataset using the proposed PSCE.}
\label{fig:CF-Path}
\end{figure*}

To supplement the introduction of the main paper and the first challenge, we consider the following example:
\begin{example}
Given an image from the MNIST dataset \citep{LeCun1998GradientbasedLA, 6296535} where a model predicts the digit $y = 9$. We aim to find a counterfactual classified as $y' = 8$ with high posterior confidence and bounded variance. Figures \ref{fig:distribution} and \ref{fig:CF-Path}  show the counterfactual generation path and the distribution of predictive probabilities on stochastic model samples. The result satisfies the $\delta$-safe and $\epsilon$-robust conditions, offering a plausible and reliable counterfactual.
\end{example}

\subsection{Complete Ablation}
\begin{figure}[!htbp]
    \centering 
    \begin{subfigure}[b]{0.5\textwidth}
        \centering
        \includegraphics[width=0.7\linewidth]{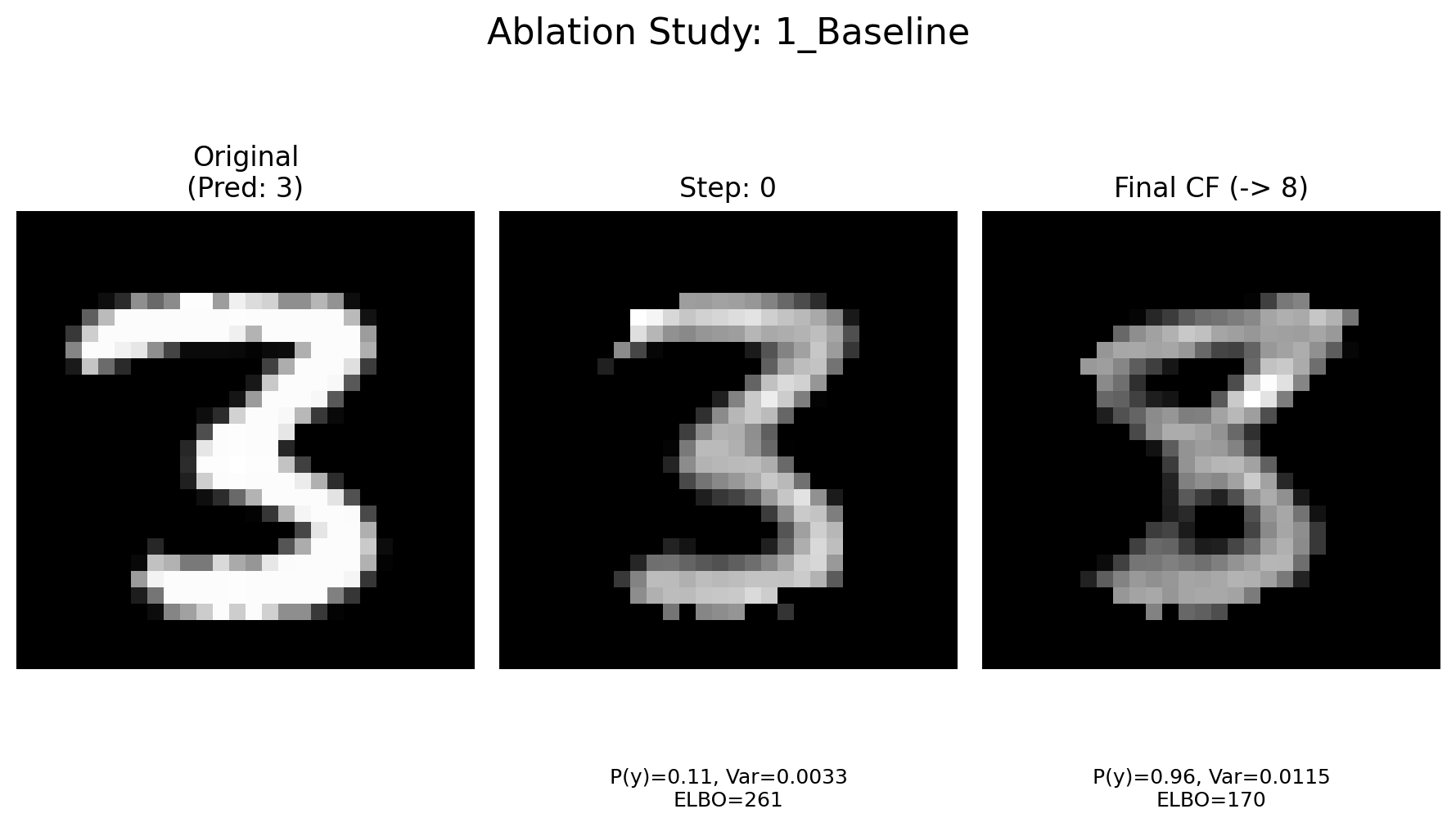}
        \caption{Combined Generation}
        \label{fig:sub_a}
    \end{subfigure}
    \begin{subfigure}[b]{0.5\textwidth}
        \centering
        \includegraphics[width=0.7\linewidth]{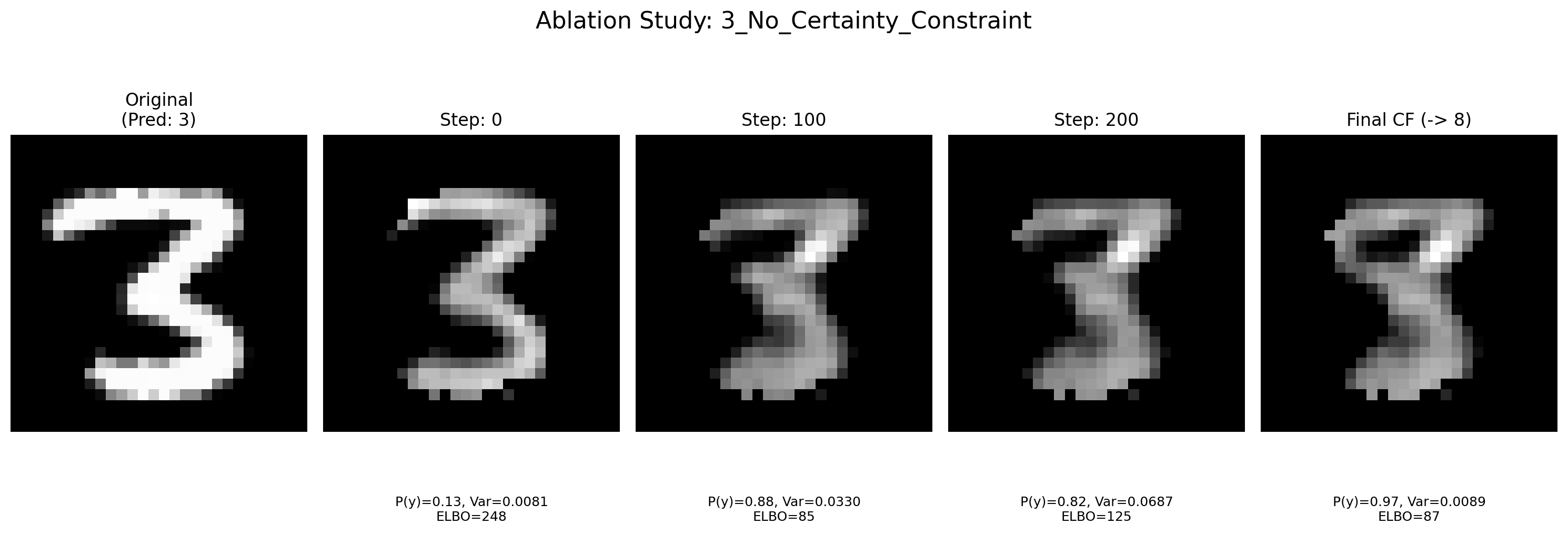}
        \caption{No $\delta$-safety}
        \label{fig:sub_b}
    \end{subfigure}
    \begin{subfigure}[b]{0.5\textwidth}
        \centering
        \includegraphics[width=0.7\linewidth]{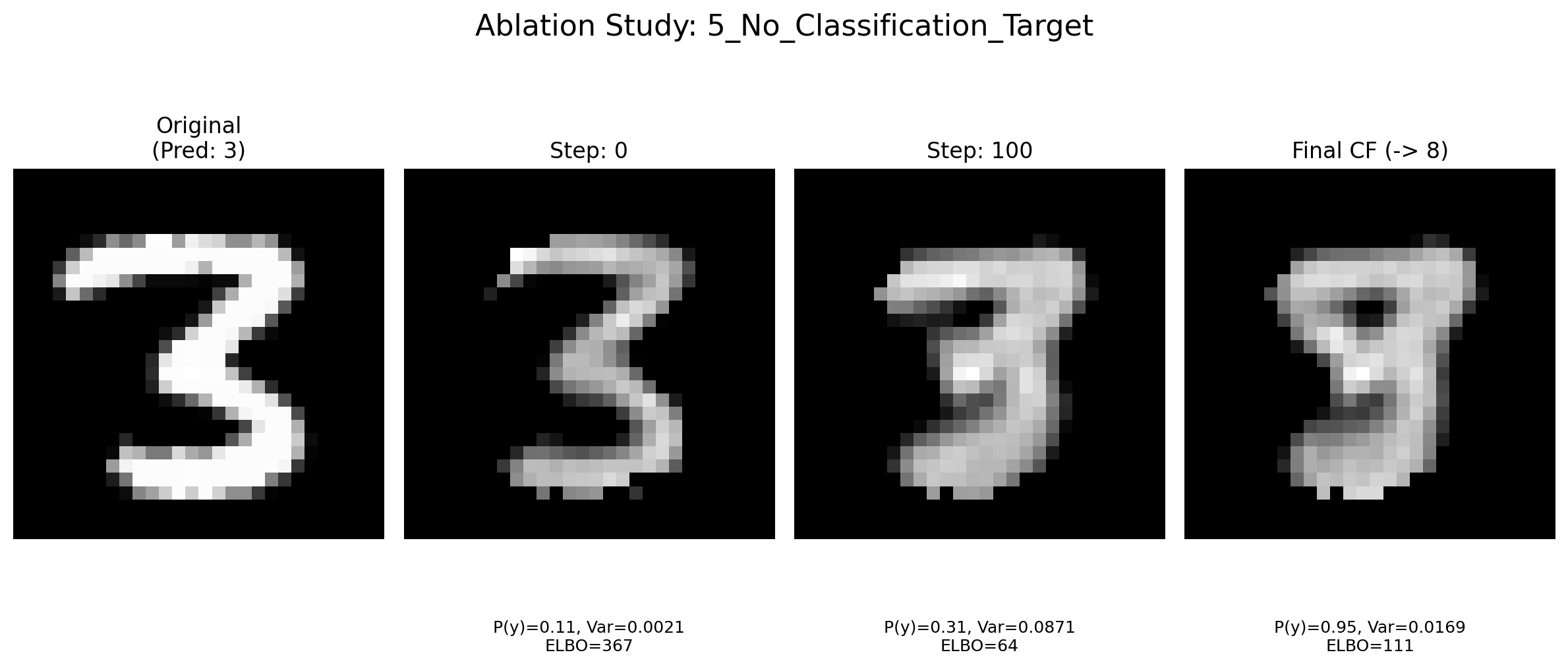}
        \caption{No cross entropy towards a desired class.}
        \label{fig:sub_c}
    \end{subfigure}

    \vspace{0.5cm}

    \begin{subfigure}[b]{0.5\textwidth}
        \centering
        \includegraphics[width=0.7\linewidth]{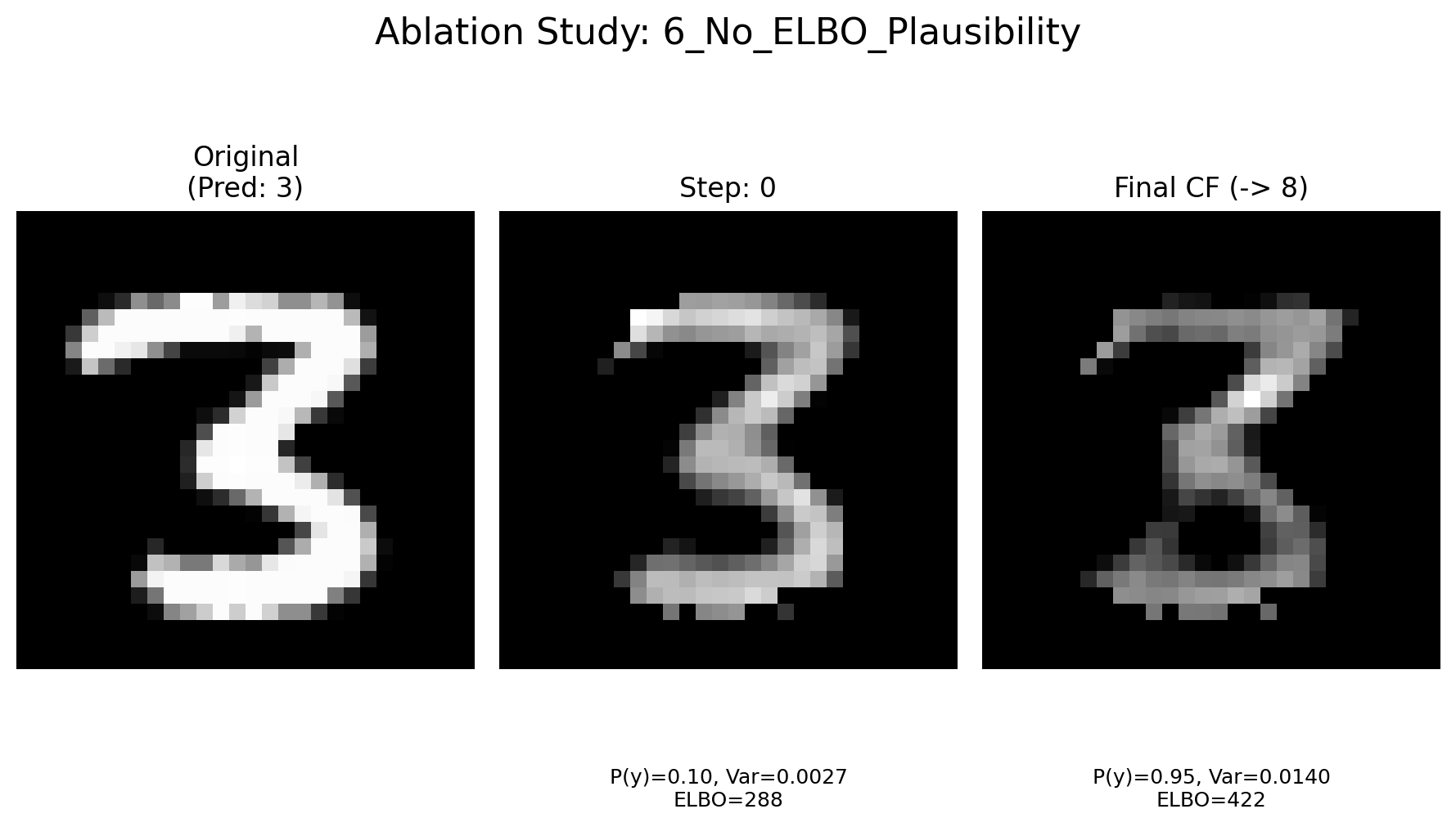}
        \caption{No ELBO}
        \label{fig:sub_d}
    \end{subfigure}
    \begin{subfigure}[b]{0.5\textwidth}
        \centering
        \includegraphics[width=0.7\linewidth]{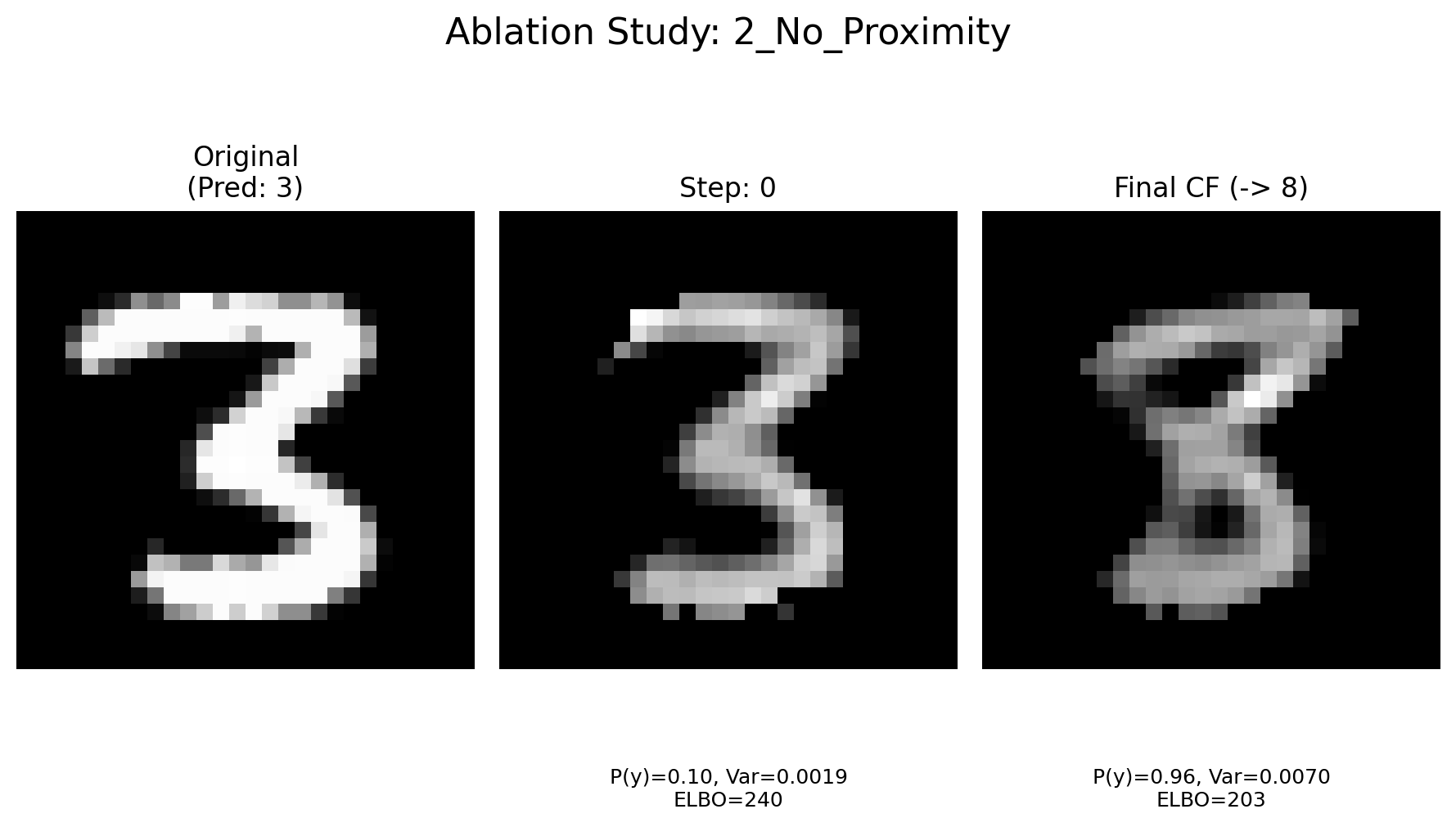}
        \caption{No latent space proximity}
        \label{fig:sub_e}
    \end{subfigure}
    \begin{subfigure}[b]{0.5\textwidth}
        \centering
        \includegraphics[width=0.7\linewidth]{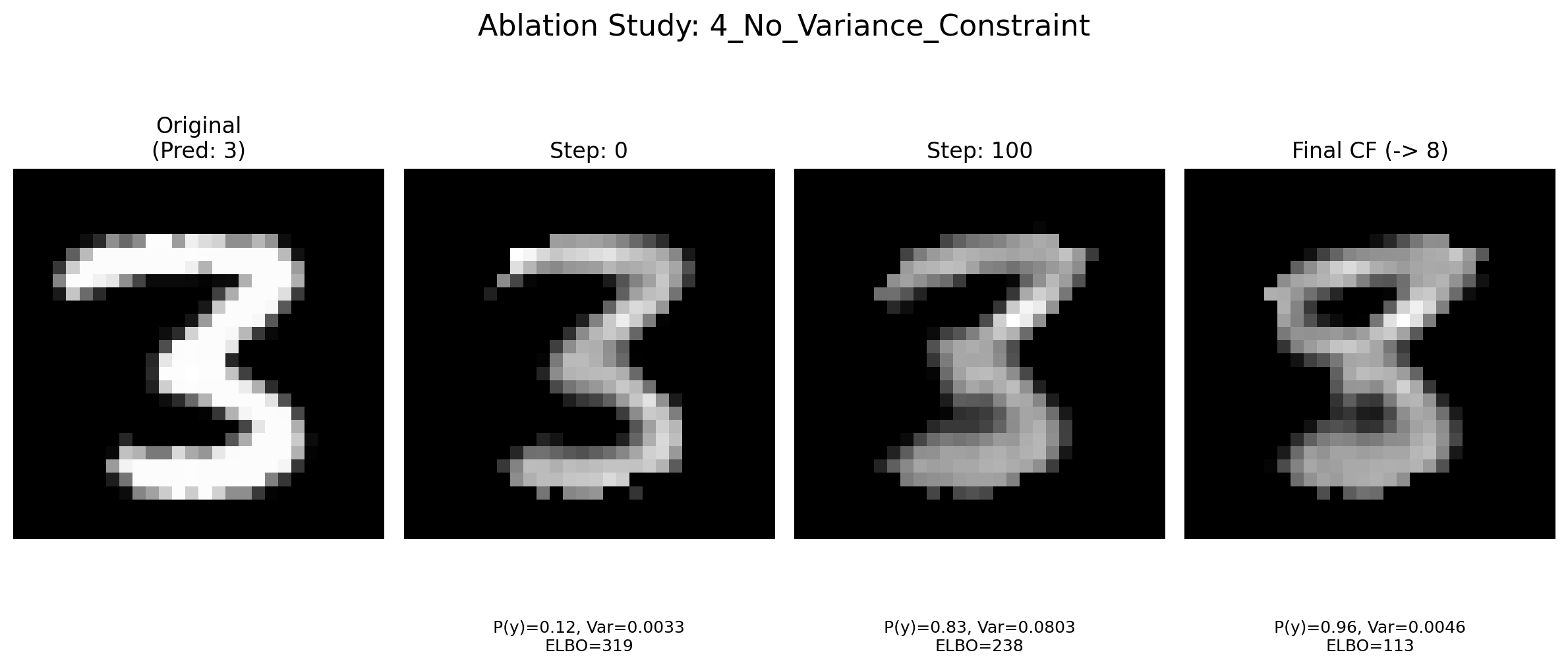}
        \caption{No $\epsilon$-robustness}
        \label{fig:sub_f}
    \end{subfigure}

    \caption{Visual effects of hyperparameter ablation.}
    \label{fig:main_figure}
\end{figure}

In this section, we show the ablation of different hyper-parameters and the associated image generated under each condition. Thus, in Figure \ref{fig:sub_a} we see the baseline image where all hyperparameters are active in the counterfactual instance generation process. Then, in Figure \ref{fig:sub_b} we remove the term encouraging $\delta$-safety; in Figure \ref{fig:sub_c} we remove the cross-entropy term that encourages the counterfactual instance to reside in the counterfactual class; Figure\ref{fig:sub_d} removes the ELBO term; Figure \ref{fig:sub_e} removes the latent distance term, and thus discouraging proximity in the latent space for the generated counterfactual instance; Figure \ref{fig:sub_f} removes the term encouraging $\epsilon$-robustness. 

\section{GENERAL DETAILS}

\subsection{Model Performance Across Seeds}
In Table \ref{tab:model_accuracy}, we provide details on the model accuracy across each of the 5 seeds on the 4 datasets experimented in the main manuscript. By which, each of these seeds the counterfactual explanation methods were deployed over 50 instances from the test dataset for each dataset.  
\begin{table}[h!] 
\centering
\caption{Model accuracy for each seed the experiments are run on. Since \textbf{MC Dropout} in the context of our work, only utilises dropout at inference time for counterfactual instance generation, we record the baseline model performance without dropout enabled.}
\label{tab:model_accuracy} 
\begin{tabular}{lcr}
\toprule 
\textbf{Dataset} & \textbf{Seed} & \textbf{Accuracy (\%)} \\
\midrule 
 & 1 & 96.49 \\
 & 2 & 95.61\\
Breast Cancer & 3 & 95.61 \\
 & 4 & 95.61 \\
 & 5 & 95.61 \\
\midrule 
 & 1 & 87.50 \\
 & 2 & 79.50 \\
Credit & 3 & 79.00  \\
 & 4 & 78.50  \\
 & 5 & 79.00 \\
\midrule 
 & 1 & 99.18 \\
 & 2 & 99.01 \\
MNIST & 3 & 99.07 \\
& 4 & 99.04 \\
 & 5 & 99.19 \\
\midrule 
 & 1 & 93.92 \\
 & 2 & 94.25 \\
Spam & 3 & 94.14 \\
 & 4 & 94.03 \\
 & 5 & 93.59 \\
 \midrule
  & 1 & 83.33  \\
 & 2 & 82.53 \\
PneumoniaMNIST & 3 & 85.10  \\
 & 4 & 87.02 \\
 & 5 & 85.26 \\
\bottomrule 
\end{tabular}
\end{table}

\begin{table}[h!] 
\centering
\caption{Model accuracy for each seed the experiments are run on for the \textbf{BNN}.}
\label{tab:model_accuracy} 
\begin{tabular}{lcr}
\toprule 
\textbf{Dataset} & \textbf{Seed} & \textbf{Accuracy (\%)} \\
\midrule 
 & 1 &  96.49 \\
 & 2 & 95.61 \\
Breast Cancer & 3 & 95.61  \\
 & 4 & 95.61 \\
 & 5 & 95.61 \\
\midrule 
 & 1 & 79.00 \\
 & 2 & 78.50 \\
Credit & 3 & 77.50 \\
 & 4 & 78.50 \\
 & 5 & 77.00 \\
\midrule 
 & 1 &  99.11 \\
 & 2 &  99.14 \\
MNIST & 3 & 98.98 \\
& 4 & 99.10 \\
 & 5 & 99.17 \\
\midrule 
 & 1 & 93.81 \\
 & 2 & 93.70 \\
Spam & 3 & 94.03 \\
 & 4 & 93.27 \\
 & 5 & 93.49  \\
 \midrule
  & 1 & 84.78 \\
 & 2 &  84.46 \\
PneumoniaMNIST & 3 & 85.58  \\
 & 4 & 84.62 \\
 & 5 & 84.78 \\
\bottomrule
\end{tabular}
\end{table}

\subsection{Hyperparmeter configuration for Counterfactual methods on each dataset}
The hyperparameters were selected subject to performance on the BNN. Each methods hyperparameters were selected based on the validity of counterfactuals. 
\begin{table}[h!]
\centering
\caption{Hyperparameters for Counterfactual Generation Methods for the Spambase and Credit datasets.}
\label{tab:cf_hyperparams}
\begin{tabular}{@{}lllc@{}}
\toprule
\textbf{Method} & \textbf{Hyperparameter} & \textbf{Description} & \textbf{Value} \\
\midrule

\multirow{8}{*}{\begin{tabular}[c]{@{}c@{}}PSCE \end{tabular}} & Optimizer &   Optimization algorithm & Adam \\
& Learning Rate & Step size for the optimizer & $0.1$ \\
& Max Iterations & Maximum optimization steps & $2000$ \\
& Proximity Weight ($\lambda_{prox}$) & Weight for the latent space distance loss & $0.001$ \\
& ELBO Weight ($\lambda_{ELBO}$) & Weight for the VAE's ELBO loss & $0.002$ \\
& Classification Weight ($\lambda_{class}$) & Weight for the negative log probability loss & $1.0$ \\
& Target Probability ($\delta$) & Minimum required probability for the target class & $\ge 0.95$ \\
& Variance Threshold ($\epsilon$) & Maximum allowed variance in predictions & $\le 0.01$ \\
\midrule

\multirow{4}{*}{\begin{tabular}[c]{@{}c@{}}BayesCF \end{tabular}} & Optimizer &   Optimization algorithm & Adam \\
& Learning Rate & Step size for the optimizer & $0.1$ \\
& Max Iterations & Maximum optimization steps & $2000$ \\
& Proximity Weight ($\lambda_{prox}$) & Weight for the L2 distance loss in input space & $0.001$ \\
\midrule

\multirow{4}{*}{Schut} & Max Iterations & Maximum number of feature perturbations & $2000$ \\
& Step Size & Magnitude of change for a perturbed feature & $0.1$ \\
& Max Feature Changes & Maximum times a single feature can be changed & $20$ \\
& Confidence Threshold & Minimum required confidence for the target class & $\ge 0.95$ \\
\bottomrule
\end{tabular}
\end{table}

\begin{table}[h!]
\centering
\caption{Hyperparameters for Counterfactual Generation Methods on the Wisconsin Breast Cancer Dataset.}
\label{tab:cf_hyperparams_v2}
\begin{tabular}{@{}lllc@{}}
\toprule
\textbf{Method} & \textbf{Hyperparameter} & \textbf{Description} & \textbf{Value} \\
\midrule

\multirow{9}{*}{\begin{tabular}[c]{@{}c@{}}PSCE\end{tabular}} & Optimizer &   Optimization algorithm & Adam \\
& Learning Rate & Step size for the optimizer & $0.1$ \\
& Max Iterations & Maximum optimization steps & $2000$ \\
& Proximity Weight ($\lambda_{prox}$) & Weight for the latent space distance loss & $0.2$ \\
& ELBO Weight ($\lambda_{ELBO}$) & Weight for the VAE's ELBO loss & $0.1$ \\
& Target Prob. Weight ($\lambda_{\delta}$) & Weight for the target probability constraint loss & $0.2$ \\
& Variance Weight ($\lambda_{\epsilon}$) & Weight for the variance constraint loss & $0.1$ \\
& Classification Weight ($\lambda_{class}$) & Weight for the negative log likelihood loss & $0.1$ \\
& Target Probability ($\delta$) & Required confidence for the target class & $\geq 0.95$ \\
& Variance Threshold ($\epsilon$) & Ideal variance in predictions & $\leq 0.01$ \\
\midrule

\multirow{4}{*}{\begin{tabular}[c]{@{}c@{}}BayesCF \end{tabular}} & Optimizer &   Optimization algorithm & Adam \\
& Learning Rate & Step size for the optimizer & $0.1$ \\
& Max Iterations & Maximum optimization steps & $2000$ \\
& Proximity Weight ($\lambda_{prox}$) & Weight for the L2 distance loss in input space & $0.1$ \\
\midrule

\multirow{4}{*}{Schut} & Max Iterations & Maximum number of feature perturbations & $2000$ \\
& Step Size & Magnitude of change for a perturbed feature & $0.1$ \\
& Max Feature Changes & Maximum times a single feature can be changed & $10$ \\
& Confidence Threshold & Ideal confidence for the target class & $\ge 0.95$ \\
\bottomrule
\end{tabular}
\end{table}

\begin{table}[h!]
\centering
\caption{Hyperparameters for Counterfactual Generation Methods on the MNIST Dataset.}
\label{tab:cf_hyperparams_v2}
\begin{tabular}{@{}lllc@{}}
\toprule
\textbf{Method} & \textbf{Hyperparameter} & \textbf{Description} & \textbf{Value} \\
\midrule

\multirow{9}{*}{\begin{tabular}[c]{@{}c@{}}PSCE\end{tabular}} & Optimizer &   Optimization algorithm & Adam \\
& Learning Rate & Step size for the optimizer & $0.1$ \\
& Max Iterations & Maximum optimization steps & $2000$ \\
& Proximity Weight ($\lambda_{prox}$) & Weight for the latent space distance loss & $0.0001$ \\
& ELBO Weight ($\lambda_{ELBO}$) & Weight for the VAE's ELBO loss &  $0.0001$\\
& Target Prob. Weight ($\lambda_{\delta}$) & Weight for the target probability constraint loss &  $1.0$ \\
& Variance Weight ($\lambda_{\epsilon}$) & Weight for the variance constraint loss & $0.1$ \\
& Classification Weight ($\lambda_{class}$) & Weight for the negative log likelihood loss & $1.0$  \\
& Target Probability ($\delta$) & Required confidence for the target class & $\geq 0.95$ \\
& Variance Threshold ($\epsilon$) & Ideal variance in predictions & $\leq 0.01$ \\
\midrule

\multirow{4}{*}{\begin{tabular}[c]{@{}c@{}}BayesCF \end{tabular}} & Optimizer &   Optimization algorithm & Adam \\
& Learning Rate & Step size for the optimizer & $0.1$ \\
& Max Iterations & Maximum optimization steps & $2000$ \\
& Proximity Weight ($\lambda_{prox}$) & Weight for the L2 distance loss in input space & $0.0000001$ \\
\midrule

\multirow{4}{*}{Schut} & Max Iterations & Maximum number of feature perturbations & $2000$ \\
& Step Size & Magnitude of change for a perturbed feature & $0.2$\\
& Max Feature Changes & Maximum times a single feature can be changed & $5$\\
& Confidence Threshold & Ideal confidence for the target class & $\ge 0.95$ \\
\bottomrule
\end{tabular}
\end{table}

\subsection{Model Architectures}
For the implementation of each network, for the MC Dropout-based approaches, VAE and AE we use PyTorch\footnote{\url{https://pytorch.org/}}. For the BNN approach we use the bnntorch\footnote{\url{https://pypi.org/project/torchbnn/}} \cite{lee2022graddiv} library. Each network is optimised using the Adaptive Moment Estimation (ADAM) \citep{Kingma2014AdamAM} optimiser. We use the Log-Softmax activation function for classification and optimize the Negative Log Likelihood (NLL).

\begin{table}[h!]
\centering
\caption{Architecture of the MC Dropout on the Spambase, Credit and Wisconsin Breast Cancer datasets.}
\label{tab:bnn_architecture}
\begin{tabular}{llll}
\toprule
\textbf{Layer} & \textbf{Type} & \textbf{Configuration} & \textbf{Activation} \\
\midrule
Input & - & Size: $D_{in}$ & - \\
1 & Fully Connected & $D_{in} \rightarrow 64$ & ReLU \\
2 & Dropout & $p=0.5$ & - \\
3 & Fully Connected & $64 \rightarrow 32$ & ReLU \\
4 & Dropout & $p=0.5$ & - \\
5 & Fully Connected & $32 \rightarrow C$ & Log-Softmax \\
\bottomrule
\end{tabular}
\end{table}

\begin{table}[h!]
\centering
\caption{Architecture of the BNN Classifier with Bayesian Layers on the Spambase, Credit and Wisconsin Breast Cancer datasets.}
\label{tab:tabular_bnn_architecture}
\begin{tabular}{llll}
\toprule
\textbf{Layer} & \textbf{Type} & \textbf{Configuration} & \textbf{Activation} \\
\midrule
Input & - & Size: $D_{in}$ & - \\
1 & Bayesian Linear & $D_{in} \rightarrow 64$ & ReLU \\
2 & Bayesian Linear & $64 \rightarrow 32$ & ReLU \\
3 & Bayesian Linear & $32 \rightarrow C$ & Log-Softmax \\
\bottomrule
\end{tabular}
\end{table}
\begin{table}[h!]
\centering
\caption{Architecture of the Variational Autoencoder (VAE) on the Spambase, Credit and Wisconsin Breast Cancer datasets.}
\label{tab:vae_architecture}
\begin{tabular}{llll}
\toprule
\textbf{Component / Layer} & \textbf{Type} & \textbf{Configuration} & \textbf{Activation} \\
\midrule
\multicolumn{4}{c}{\textit{\textbf{Encoder}}} \\
\midrule
Input & - & Size: $D_{in}$ & - \\
Encoder FC1 & Fully Connected & $D_{in} \rightarrow 40$ & ReLU \\
Mean ($\mu$) & Fully Connected & $40 \rightarrow D_{z}$ & Linear \\
Log-Variance ($\log\sigma^2$) & Fully Connected & $40 \rightarrow D_{z}$ & Linear \\
\midrule
\multicolumn{4}{c}{\textit{\textbf{Decoder}}} \\
\midrule
Input (Latent) & Reparameterized $z$ & Size: $D_{z}$ & - \\
Decoder FC1 & Fully Connected & $D_{z} \rightarrow 40$ & ReLU \\
Output & Fully Connected & $40 \rightarrow D_{in}$ & Linear \\
\bottomrule
\end{tabular}
\end{table}
\begin{table}[h!]
\centering
\caption{Architecture of the Autoencoder (AE) for the IM1 metric on the Spambase, Credit and Wisconsin Breast Cancer datasets.}
\label{tab:ae_architecture}
\begin{tabular}{llll}
\toprule
\textbf{Component / Layer} & \textbf{Type} & \textbf{Configuration} & \textbf{Activation} \\
\midrule
\multicolumn{4}{c}{\textit{\textbf{Encoder}}} \\
\midrule
Input & - & Size: $D_{in}$ & - \\
Encoder FC1 & Fully Connected & $D_{in} \rightarrow 40$ & ReLU \\
Encoder FC2 & Fully Connected & $40 \rightarrow D_{z}$ & ReLU \\
\midrule
\multicolumn{4}{c}{\textit{\textbf{Decoder}}} \\
\midrule
Input (Latent) & Encoded Vector & Size: $D_{z}$ & - \\
Decoder FC1 & Fully Connected & $D_{z} \rightarrow 40$ & ReLU \\
Output & Fully Connected & $40 \rightarrow D_{in}$ & Linear \\
\bottomrule
\end{tabular}
\end{table}

\begin{table}[h!]
\centering
\caption{Architecture of the CNN Classifier for MNIST used for MC Dropout.}
\label{tab:mnist_cnn_architecture}
\begin{tabular}{llll}
\toprule
\textbf{Layer} & \textbf{Type} & \textbf{Configuration} & \textbf{Layer Output Shape} \\
\midrule
Input & - & - & (1, 28, 28) \\
\midrule
Conv-1 & Conv2D & 10 filters, kernel=5 & (10, 24, 24) \\
Pool-1 & Max Pooling & kernel=2, stride=2 & (10, 12, 12) \\
\multicolumn{4}{l}{\textit{Activation: ReLU}} \\
\midrule
Conv-2 & Conv2D & 20 filters, kernel=5 & (20, 8, 8) \\
Drop-2 & Dropout2D & p=0.5 (default) & (20, 8, 8) \\
Pool-2 & Max Pooling & kernel=2, stride=2 & (20, 4, 4) \\
\multicolumn{4}{l}{\textit{Activation: ReLU}} \\
\midrule
Flatten & Flatten & - & (320) \\
\midrule
FC-1 & Fully Connected & 320 $\rightarrow$ 50 & (50) \\
\multicolumn{4}{l}{\textit{Activation: ReLU}} \\
Drop-3 & Dropout & p=0.5 (default) & (50) \\
\midrule
FC-2 (Output) & Fully Connected & 50 $\rightarrow$ 10 & (10) \\
\multicolumn{4}{l}{\textit{Activation: Log-Softmax}} \\
\bottomrule
\end{tabular}
\end{table}

\begin{table}[h!]
\centering
\caption{Architecture of the Variational Autoencoder (VAE) for the MNIST datasets.}
\label{tab:mnist_vae_architecture}
\begin{tabular}{llll}
\toprule
\textbf{Component / Layer} & \textbf{Type} & \textbf{Configuration} & \textbf{Activation} \\
\midrule
\multicolumn{4}{c}{\textit{\textbf{Encoder}}} \\
\midrule
Input & - & Flattened Image (784) & - \\
FC1 & Fully Connected & 784 $\rightarrow$ 400 & ReLU \\
Mean ($\mu$) & Fully Connected & 400 $\rightarrow$ 20 & Linear \\
Log-Var ($\log\sigma^2$) & Fully Connected & 400 $\rightarrow$ 20 & Linear \\
\midrule
\multicolumn{4}{c}{\textit{\textbf{Decoder}}} \\
\midrule
Input (Latent) & Reparameterized $z$ & Size: 20 & - \\
FC3 & Fully Connected & 20 $\rightarrow$ 400 & ReLU \\
FC4 (Output) & Fully Connected & 400 $\rightarrow$ 784 & ReLU \\
\bottomrule
\end{tabular}
\end{table}

\begin{table}[h!]
\centering
\caption{Architecture of the Autoencoder (AE) for the MNIST datasets.}
\label{tab:mnist_ae_architecture}
\begin{tabular}{llll}
\toprule
\textbf{Component / Layer} & \textbf{Type} & \textbf{Configuration} & \textbf{Activation} \\
\midrule
\multicolumn{4}{c}{\textit{\textbf{Encoder}}} \\
\midrule
Input & - & Flattened Image (784) & - \\
Encoder FC1 & Fully Connected & 784 $\rightarrow$ 200 & ReLU \\
Encoder FC2 & Fully Connected & 200 $\rightarrow$ 20 & ReLU \\
\midrule
\multicolumn{4}{c}{\textit{\textbf{Decoder}}} \\
\midrule
Input (Latent) & Encoded Vector & Size: 20 & - \\
Decoder FC1 & Fully Connected & 20 $\rightarrow$ 200 & ReLU \\
Decoder FC2 (Output) & Fully Connected & 200 $\rightarrow$ 784 & Linear \\
\bottomrule
\end{tabular}
\end{table}

\begin{table}[h!]
\centering
\caption{Architecture of the BNN-CNN with Bayesian Layers for the MNIST datasets.}
\label{tab:bnn_cnn_architecture}
\begin{tabular}{llll}
\toprule
\textbf{Layer} & \textbf{Type} & \textbf{Configuration} & \textbf{Layer Output Shape} \\
\midrule
Input & - & - & (1, 28, 28) \\
\midrule
Conv-1 & Bayesian Conv2D & 10 filters, kernel=5 & (10, 24, 24) \\
Pool-1 & Max Pooling & kernel=2, stride=2 & (10, 12, 12) \\
\multicolumn{4}{l}{\textit{Activation: ReLU}} \\
\midrule
Conv-2 & Bayesian Conv2D & 20 filters, kernel=5 & (20, 8, 8) \\
Pool-2 & Max Pooling & kernel=2, stride=2 & (20, 4, 4) \\
\multicolumn{4}{l}{\textit{Activation: ReLU}} \\
\midrule
Flatten & Flatten & - & (320) \\
\midrule
FC-1 & Bayesian Linear & 320 $\rightarrow$ 50 & (50) \\
\multicolumn{4}{l}{\textit{Activation: ReLU}} \\
\midrule
FC-2 (Output) & Bayesian Linear & 50 $\rightarrow$ 10 & (10) \\
\multicolumn{4}{l}{\textit{Activation: Log-Softmax}} \\
\bottomrule
\end{tabular}
\end{table}

\section{Metrics for CE Evaluation}
\subsection{IM1} 
The IM1 metric \citep{10.1007/978-3-030-86520-7_40, Schut2021GeneratingIC}, evaluates the ratio of the reconstruction loss with respect to an autoencoder trained on the counterfactual class $y^\prime$ and the reconstruction loss of a counterfactual with an autoencoder (AE) trained on the original class $y$. 
Let $\text{AE}_y$ be the autoencoder trained on the original class, and $\text{AE}_{y^\prime}$ be the autoencoder trained on the counterfactual class. Then, we have: 
\begin{align*}
    \text{IM1}(\mathbf{x}^\prime) = \frac{\Vert \mathbf{x}^\prime - \text{AE}_{y^\prime}(\mathbf{x}^\prime)\Vert^2_2}{\Vert \mathbf{x}^\prime - \text{AE}_y(\mathbf{x}^\prime) \Vert^2_2}.
\end{align*}
\subsection{Robustness Ratio} 

The robustness ratio adopted in \citep{batten2025uncertaintyaware}, explores the robustness of counterfactual explanations subject to a small perturbation in the input instance. More formally: 
\begin{align*}
    \text{RR}(\mathbf{x}^\prime;\kappa) = \frac{\Vert \mathcal{G}_{CF}(\mathbf{x} + \kappa) - \mathbf{x}^\prime  \Vert_2^2}{\Vert\mathbf{x}^\prime - \mathbf{x} \Vert^2_2 }
\end{align*}
for some $\kappa << 1$. We follow implementation details of \citep{batten2025uncertaintyaware} and set $\kappa = 1e-3$. 

\subsection{Implausibility}
Implausibility \citep{10.1609/aaai.v38i10.28956} evaluates the mean distance to other instances of the counterfactual class within the training dataset. More formally: 
\begin{align*}
    \text{Imp}(\mathbf{x}^\prime) = \frac{1}{\vert \mathbf{X}_{y^\prime}\vert} \sum_{\mathbf{x} \in \mathbf{X}_{y^\prime}} \text{dist}(\mathbf{x}, \mathbf{x}^\prime),
\end{align*}
where $\mathbf{X}_{y^\prime} \subset \mathbf{X}$ denotes a subset of training data for the target counterfactual class $y^\prime$, and $\mathbf{x}^\prime$ is a counterfactual instance for $\mathbf{x} \in \mathbf{X}_{y^\prime}$. 

\subsection{Validity} 
Validity \citep{10.1145/3351095.3372850, batten2025uncertaintyaware} looks at the percentage of correctly classified counterfactuals, that is how many counterfactuals belong to the counterfactual class. More formally: 
\begin{align*}
\text{Validity}(\mathbf{X}')
&= \frac{1}{\lvert \mathbf{X}^\prime \rvert}
\sum_{\mathbf{x}' \in \mathbf{X}'}
\mathds{1}\!\left[
\underset{c \in \{1,\ldots,C\}}{\operatorname{arg\,max}}
\ p(y = c \mid \mathbf{x}^\prime, \mathcal{D})
= y^\prime
\right].
\end{align*}
returning the fraction of counterfactual instances which are correctly classified as $y^\prime$ from the set of generated counterfactuals $X^\prime$.

\section{PSCE Examples on MNIST}
\begin{figure}
    \centering
    \includegraphics[width=1\linewidth]{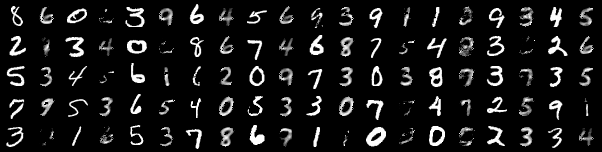}
    \caption{Generated counterfactual examples on a single seed of the BNN.}
    \label{fig:all_examples}
\end{figure}

From Figure \ref{fig:all_examples} we present examples of counterfactuals generated by our method. 
\begin{figure}
    \centering
    \includegraphics[width=1\linewidth]{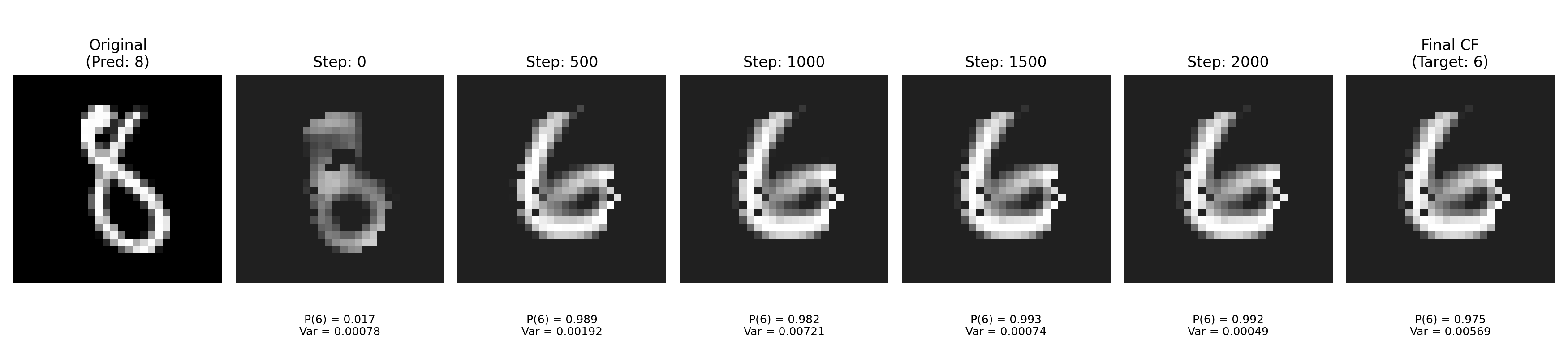}

    \label{fig:placeholder}
\end{figure}
\begin{figure}
    \centering
    \includegraphics[width=1\linewidth]{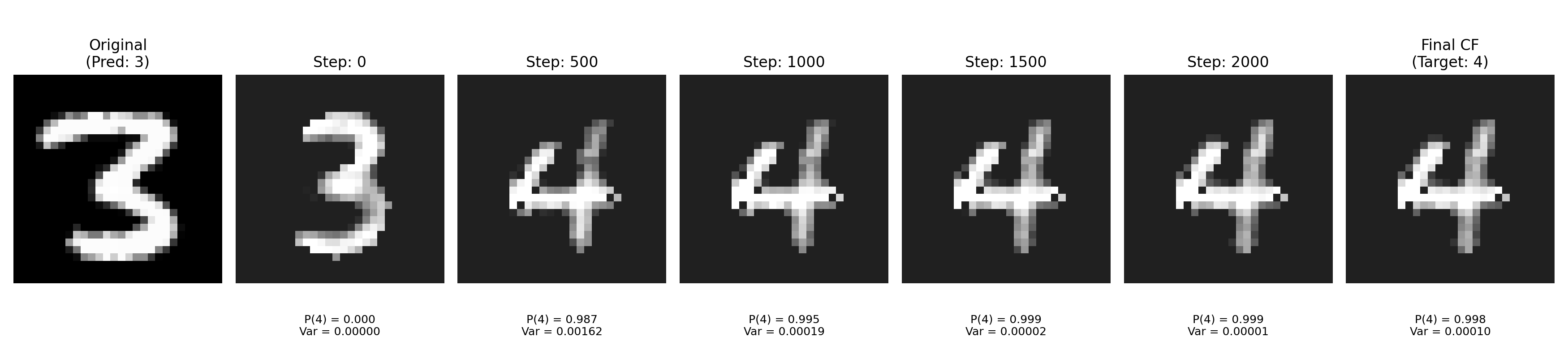}

    \label{fig:placeholder}
\end{figure}
\begin{figure}
    \centering
    \includegraphics[width=1\linewidth]{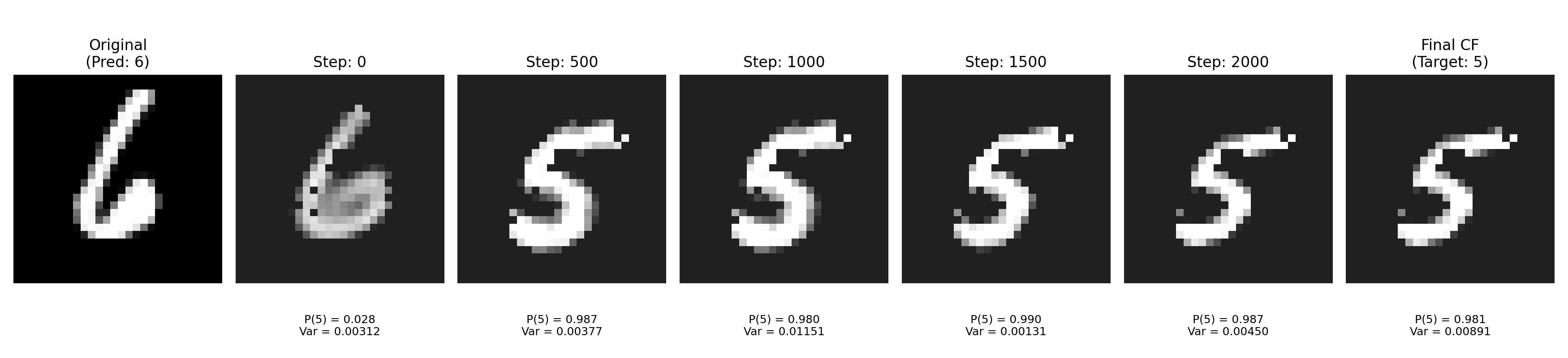}

    \label{fig:placeholder}
\end{figure}

\begin{figure}
    \centering
    \includegraphics[width=1\linewidth]{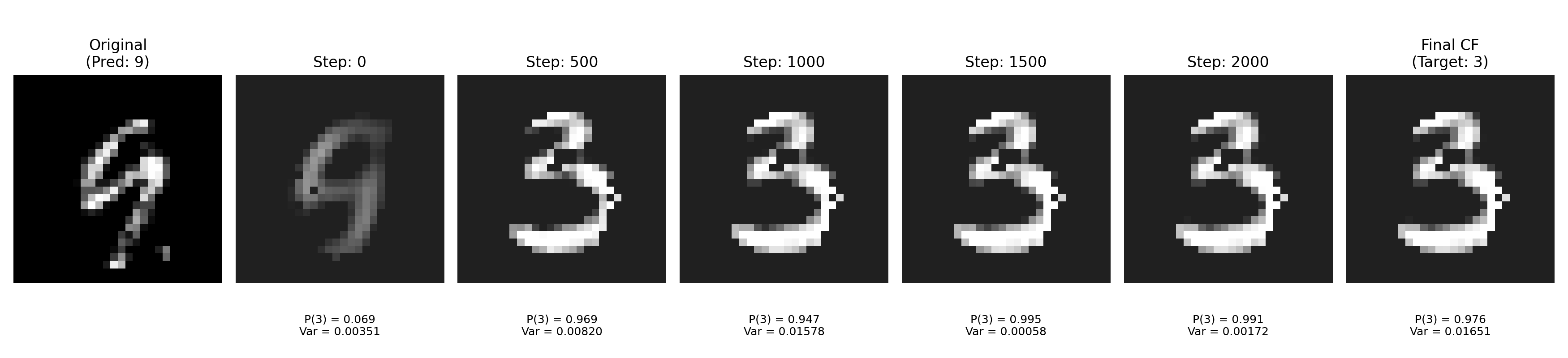}

    \label{fig:placeholder}
\end{figure}

\begin{figure}
    \centering
    \includegraphics[width=1\linewidth]{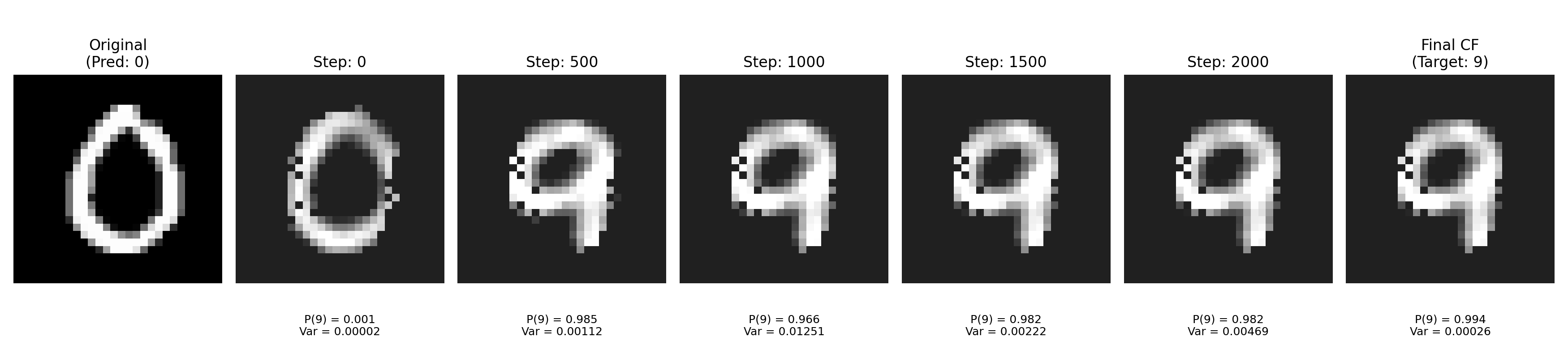}

    \label{fig:placeholder}
\end{figure}
\begin{figure}
    \centering
    \includegraphics[width=1\linewidth]{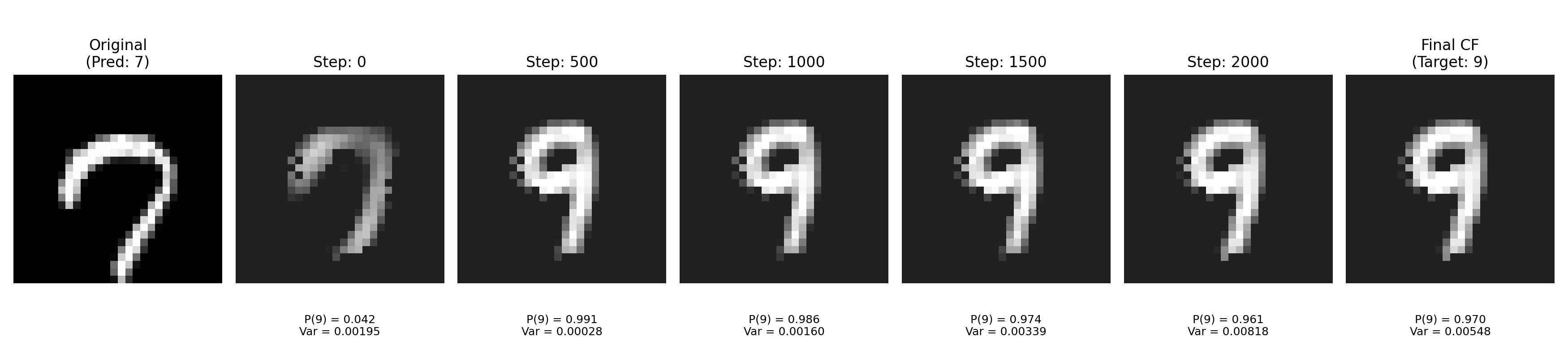}

    \label{fig:placeholder}
\end{figure}
    \begin{figure}
    \centering
    \includegraphics[width=1\linewidth]{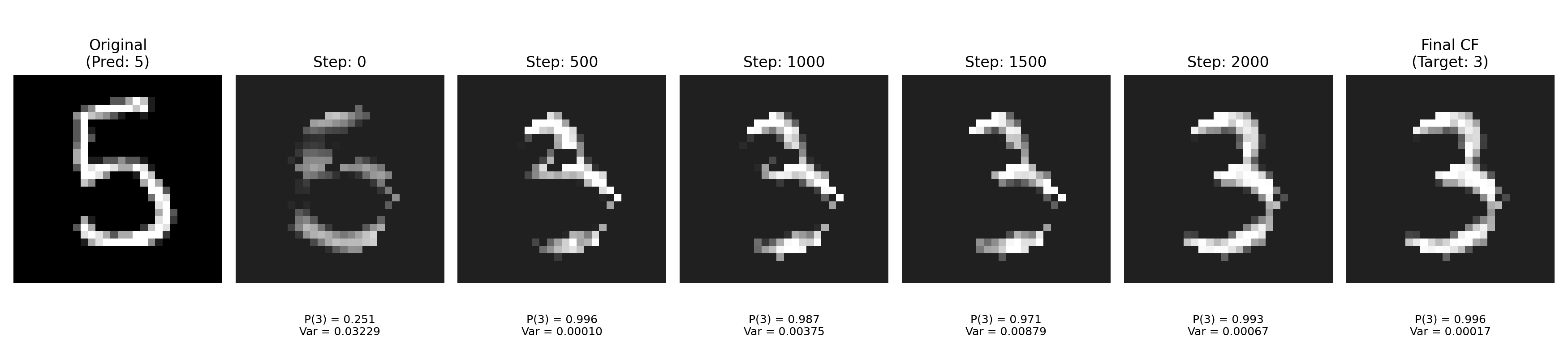}

    \label{fig:placeholder}
\end{figure}    \begin{figure}
    \centering
    \includegraphics[width=1\linewidth]{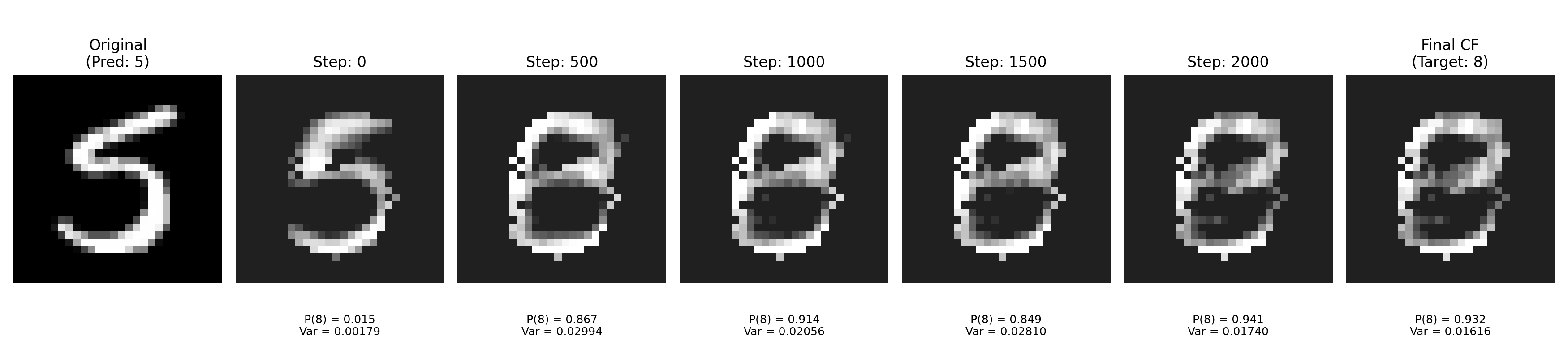}

    \label{fig:placeholder}
\end{figure}

\section{Additional Comparison with Non-Bayesian Baselines}

For clarity in the main article we restricted the results including non-Bayeisan baselines outlined in Tables \ref{upd_tb1} and \ref{upd_tb2}, where we consider the Wachter's approach \citep{wachter} and RObust Algorithmic Recourse (ROAR) \citep{upadhyay2021towards} following their optimisation procedure.  

\begin{table*}[ht]
\centering
\caption{Performance metrics measured across all datasets. The best performing method is highlighed in \textbf{bold} text. The experiments are run over 5 runs and the \emph{mean} and \emph{standard deviation} are recorded. Each run is evaluated over 100 instances from the test data. These results use \textbf{BNN}.}
\label{tab:flipped_results}
\begin{adjustbox}{width=\textwidth}
\begin{tabular}{llccccc}
\toprule
\textbf{Dataset} & \textbf{Metric} & \textbf{PSCE (Ours)} & \textbf{BayesCF} & \textbf{Schut} & \textbf{Wachter} & \textbf{ROAR} \\
\midrule

\multirow{4}{*}{Credit}
  & IM1 $(\downarrow)$              & \textbf{0.6561 $\pm$ 0.0603}  &  0.8311 $\pm$ 0.0289 &  0.9042 $\pm$ 0.0222 & 1.0423 $\pm$ 0.0244 & 1.0285 $\pm$ 0.0239 \\
  & Implausibility $(\downarrow)$   & \textbf{8.4606 $\pm$ 0.2683}  & 10.1530 $\pm$ 0.3799  & 10.8363 $\pm$ 0.1218 & 9.3963 $\pm$ 0.0713 & 9.3061 $\pm$ 0.0677
 \\
  & Robustness Ratio $1e\text{-}3$ $(\downarrow)$ & \textbf{0.1400 $\pm$ 0.0203} & 0.1461 $\pm$ 0.0168  & 0.2318 $\pm$ 0.0304 & 0.5044 $\pm$ 0.0176 &   0.4863 $\pm$ 0.0458\\
  & Validity \% $(\uparrow)$        & \textbf{100.0 $\pm$ 0.0} &  \textbf{100.0 $\pm$ 0.0} & \textbf{100.0 $\pm$ 0.0} & 94.4$\pm$2.7 & 95.4$\pm$1.0\\
\midrule

\multirow{4}{*}{Breast Cancer}
  & IM1 $(\downarrow)$              & 1.5540 $\pm$ 0.2427  &  1.1957 $\pm$ 0.0507 & 1.1151 $\pm$ 0.0362 & 0.8348 $\pm$ 0.0251 & \textbf{0.8164 $\pm$ 0.0267} \\
  & Implausibility $(\downarrow)$   & \textbf{5.7579 $\pm$ 0.0439} &  7.2014 $\pm$ 0.0423 & 7.4072 $\pm$ 0.0333 & 6.9434 $\pm$ 0.0436 & 6.9247 $\pm$ 0.0443 \\
  & Robustness Ratio $1e\text{-}3$ $(\downarrow)$ & \textbf{ 0.1566 $\pm$ 0.0152} & 0.3867 $\pm$ 0.0306  &  0.3050 $\pm$ 0.0211 & 0.2684 $\pm$ 0.0205 & 0.2601 $\pm$ 0.0214\\
  & Validity \% $(\uparrow)$        & \textbf{86.8$\pm$3.2}  & 78.4$\pm$3.5  & 84.8$\pm$1.9 & 99.4$\pm$0.5 & 99.4$\pm$0.8\\
\midrule
\multirow{4}{*}{MNIST}
  & IM1 $(\downarrow)$              & \textbf{0.9768 $\pm$ 0.1598}   &  1.6408 $\pm$ 0.1267 &  1.2298 $\pm$ 0.0526 & 1.5153 $\pm$ 0.0699 & 1.4978 $\pm$ 0.0739 \\
  & Implausibility $(\downarrow)$   &  \textbf{28.1567 $\pm$ 0.2665}& 35.7763 $\pm$ 0.0737  &  31.2247 $\pm$ 0.2067 & 32.1812 $\pm$ 0.1574 & 31.8525 $\pm$ 0.1292\\
  & Robustness Ratio $1e\text{-}3$ $(\downarrow)$ & \textbf{0.7822 $\pm$ 0.0289} &  1.0191 $\pm$ 0.0463 &  0.9011 $\pm$ 0.0170 & 1.0016 $\pm$ 0.0176 & 0.8750 $\pm$ 0.0182 \\
  & Validity \% $(\uparrow)$        &  \textbf{98.2$\pm$1.6}&  67.6$\pm$5.4  &  96.0$\pm$1.8 &  47.0$\pm$5.1 &  52.8$\pm$3.7 \\
\midrule

\multirow{4}{*}{Spam}
  & IM1 $(\downarrow)$              & \textbf{0.7387 $\pm$ 0.0578}  &  1.0948 $\pm$ 0.0218 & 0.9137 $\pm$ 0.0299 & 1.1884 $\pm$ 0.0173 & 1.1588 $\pm$ 0.0160\\
  & Implausibility $(\downarrow)$   &   \textbf{ 6.1781 $\pm$ 0.0312} & 9.4032 $\pm$ 0.0471  & 12.9145 $\pm$ 0.1061 & 9.4423 $\pm$ 0.0277 & 9.4445 $\pm$ 0.0439\\
  & Robustness Ratio $1e\text{-}3$ $(\downarrow)$ & \textbf{0.1028 $\pm$ 0.0099}  & 0.1507 $\pm$ 0.0124  & 0.2046 $\pm$ 0.0083 & 0.3883 $\pm$ 0.0101 &0.3616 $\pm$ 0.0106 \\
  & Validity \% $(\uparrow)$        &  96.0 $\pm$ 3.3 & \textbf{100.0 $\pm$ 0.0}  & 98.6$\pm$0.5 & 97.0$\pm$1.1 & 98.2$\pm$0.4 \\
  \bottomrule
  \multirow{4}{*}{PneumoniaMNIST}
  & IM1 $(\downarrow)$              & \textbf{0.630 $\pm$ 0.095} & 1.062 $\pm$ 0.009  & 1.157 $\pm$ 0.024 & 1.074 $\pm$ 0.013 & 1.155 $\pm$ 0.014 \\
  & Implausibility $(\downarrow)$   & \textbf{4.465 $\pm$ 0.332} &  5.687 $\pm$ 0.027 & 5.350 $\pm$ 0.019 & 5.624 $\pm$ 0.021 & 5.273 $\pm$ 0.017 \\
  & Robustness Ratio $1e\text{-}3$ $(\downarrow)$ & \textbf{0.255 $\pm$ 0.020}  &  1.191 $\pm$ 0.009  &  1.015 $\pm$ 0.033 & 1.191 $\pm$ 0.017 & 1.079 $\pm$ 0.018\\
  & Validity \% $(\uparrow)$        & \textbf{99.8$\pm$0.4}    & 98.4$\pm$1.5  & 91.2$\pm$3.6 & 98.6$\pm$0.8 & 88.0$\pm$2.6 \\
\bottomrule
\end{tabular}\label{upd_tb1}
\end{adjustbox}
\end{table*}

\begin{table*}[ht]
\label{upd_tb2}
\centering
\caption{Performance metrics measured across all datasets. The best performing method is highlighed in \textbf{bold} text. The experiments are run over 5 runs and the \emph{mean} and \emph{standard deviation} are recorded. Each run is evaluated over 100 instances from the test data. These results use \textbf{MC Dropout}.}
\label{tab:flipped_results2}
\begin{adjustbox}{width=\textwidth}
\begin{tabular}{llccccc}
\toprule
\textbf{Dataset} & \textbf{Metric} & \textbf{PSCE (Ours)} & \textbf{BayesCF} & \textbf{Schut} & \textbf{Wachter} & \textbf{ROAR} \\
\midrule

\multirow{4}{*}{Credit}
  & IM1 $(\downarrow)$              &  \textbf{0.7383 $\pm$ 0.0341} & 1.0767 $\pm$ 0.0214      & 0.9251 $\pm$ 0.0320 & 1.0429 $\pm$ 0.0236 & 1.0275 $\pm$ 0.0261\\
  & Implausibility $(\downarrow)$   &  \textbf{8.6113 $\pm$ 0.1494}&  9.3615 $\pm$ 0.0543  & 9.9943 $\pm$ 0.2124 & 9.3773 $\pm$ 0.0528 & 9.2826 $\pm$ 0.0600\\
  & Robustness Ratio $1e\text{-}3$ $(\downarrow)$ &  \textbf{0.0622 $\pm$ 0.0103}& 0.1219 $\pm$ 0.0106  &  0.1418 $\pm$ 0.0208 & 0.4867 $\pm$ 0.0175 & 0.4951 $\pm$ 0.0304 \\
  & Validity \% $(\uparrow)$        & \textbf{100.0$\pm$0.0}  & 66.8$\pm$ 5.2  &  \textbf{100.0$\pm$0.0} & 95.2$\pm$2.2 & 96.2$\pm$1.6 \\
\midrule

\multirow{4}{*}{Breast Cancer}
  & IM1 $(\downarrow)$              & \textbf{0.6932 $\pm$ 0.0911}  &  1.1300 $\pm$ 0.0306 &  1.0332 $\pm$ 0.0340 & 0.8214 $\pm$ 0.0288 & 0.8146 $\pm$ 0.0328 \\
  & Implausibility $(\downarrow)$   & \textbf{5.6275 $\pm$ 0.0627} &  7.1505 $\pm$ 0.0323 & 7.2913 $\pm$ 0.0354 & 6.8754 $\pm$ 0.0680 & 6.8794 $\pm$ 0.0604\\
  & Robustness Ratio $1e\text{-}3$ $(\downarrow)$ & 0.0800 $\pm$ 0.0047  & \textbf{0.0539 $\pm$ 0.0104}  &  0.1045 $\pm$ 0.0105 & 0.2757 $\pm$ 0.0185 & 0.2757 $\pm$ 0.0241 \\
  & Validity \% $(\uparrow)$        &  \textbf{97.2$\pm$1.6} &  79.2$\pm$5.0 & 90.4$\pm$0.8 & 99.0$\pm$1.3 & 99.4$\pm$0.5 \\
\midrule

\multirow{4}{*}{MNIST}
  & IM1 $(\downarrow)$              &  \textbf{0.9273 $\pm$ 0.0172}  & 1.6426 $\pm$ 0.2774 & 1.2373 $\pm$ 0.0392 & 1.4984 $\pm$ 0.0636 & 1.4739 $\pm$ 0.0617 \\
  & Implausibility $(\downarrow)$   & \textbf{27.0337 $\pm$ 0.4335} & 32.9022 $\pm$ 0.2163  &  31.3665 $\pm$ 0.1397 & 31.8151 $\pm$ 0.3558 & 31.5127 $\pm$ 0.3418 \\
  & Robustness Ratio $1e\text{-}3$ $(\downarrow)$ & \textbf{0.6493 $\pm$ 0.0496}  &  0.9918 $\pm$ 0.0471  &  1.0696 $\pm$ 0.0412 & 1.0421 $\pm$ 0.0324 & 0.8721 $\pm$ 0.0297 \\
  & Validity \% $(\uparrow)$        &  99.6$\pm$0.8 & 47.8$\pm$3.3  & \textbf{100.0 $\pm$ 0.0} & 61.6$\pm$4.5 & 75.2$\pm$5.2 \\
\midrule
\multirow{4}{*}{Spam}
  & IM1 $(\downarrow)$              &   \textbf{0.8967 $\pm$ 0.0316}  & 1.4798 $\pm$ 0.0122    &  1.2756 $\pm$ 0.0225 & 1.1112 $\pm$ 0.0718 & 1.0926 $\pm$ 0.0686 \\
  & Implausibility $(\downarrow)$   & \textbf{7.2057 $\pm$ 0.0618}  &   9.4627 $\pm$ 0.0442  &  10.0120 $\pm$ 0.0461 & 9.4063 $\pm$ 0.0565 &  9.3678 $\pm$ 0.0992\\
  & Robustness Ratio $1e\text{-}3$ $(\downarrow)$ & \textbf{0.0429 $\pm$ 0.0063} &  0.0868 $\pm$ 0.0092    &    0.0870 $\pm$ 0.0151 & 0.4330 $\pm$ 0.0559 & 0.4356 $\pm$ 0.0655\\
  & Validity \% $(\uparrow)$        &  91.8 $\pm$ 4.5 & 79.0$\pm$4.0  &  \textbf{99.0$\pm$0.0} & 96.8$\pm$2.3 & 97.0$\pm$1.6 \\
  \midrule
  \multirow{4}{*}{PneumoniaMNIST}
  & IM1 $(\downarrow)$              &  \textbf{0.545 $\pm$ 0.102}  &  1.062 $\pm$ 0.012 &  1.031 $\pm$ 0.007 & 1.131 $\pm$ 0.028 & 1.109 $\pm$ 0.026 \\
  & Implausibility $(\downarrow)$   & \textbf{4.489 $\pm$ 0.149}  & 5.647 $\pm$ 0.037  & 6.066 $\pm$ 0.138 &  5.272 $\pm$ 0.024 & 5.312 $\pm$ 0.030 \\
  & Robustness Ratio $1e\text{-}3$ $(\downarrow)$ &  \textbf{0.098 $\pm$ 0.017} &  0.597 $\pm$ 0.030 & 0.770 $\pm$ 0.034 & 1.299 $\pm$ 0.023 & 1.346 $\pm$ 0.030 \\
  & Validity \% $(\uparrow)$        & 99.4$\pm$0.5 & \textbf{100.0$\pm$0.0} & \textbf{100.0$\pm$0.0} & 98.0$\pm$0.6 & 98.4$\pm$1.2 \\
\bottomrule
\end{tabular}\label{upd_tb2}
\end{adjustbox}
\end{table*}

\section{Limitation and Future work}
From a theoretical perspective, an ideal avenue would be to explicitly explore the proposed method empirically under theoretical guarantees in the context of online learning or data drift, as it would present a natural progression for the proposed methodology.

\end{document}